%% file: main.tex
\documentclass[10pt,twocolumn,letterpaper]{article}

\usepackage{cvpr}      
\usepackage{algorithm}
\usepackage{algorithmic}
\usepackage{xcolor}
\usepackage{amsmath}
\usepackage{amssymb}
\usepackage{booktabs}
\usepackage{tabularx}
\usepackage{wasysym}
\usepackage{multirow}

\definecolor{skyblue}{RGB}{0, 176, 240}
\definecolor{orange}{RGB}{234, 99, 51}
\definecolor{neon}{RGB}{128, 245, 16}
\definecolor{violet}{RGB}{102, 43, 148} 
\definecolor{greeen}{RGB}{112, 173, 71}
\definecolor{bluee}{RGB}{83, 112, 255}

\input{preamble}

\definecolor{cvprblue}{rgb}{0.21,0.49,0.74}
\usepackage[pagebackref,breaklinks,colorlinks,citecolor=cvprblue]{hyperref}

\def\paperID{*****} 
\def\confName{CVPR}
\def\confYear{2024}

\title{Semi-supervised Open-World Object Detection}

\author{Sahal Shaji Mullappilly$^{1}$ \quad
Abhishek Singh Gehlot$^{1}$ \quad
Rao Muhammad Anwer$^{1}$ \\
Fahad Shahbaz Khan$^{1,2}$ \quad
Hisham Cholakkal$^{1}$
\vspace{0.1em} \\
{\tt\small \{sahal.mullappilly,abhishek.gehlot,rao.anwer,fahad.khan,hisham.cholakkal\}@mbzuai.ac.ae}
\vspace{0.1em} \\
$^{1}$Mohamed bin Zayed University of Artificial Intelligence  $^{2}$Linköping University
}

\begin{document}
\maketitle
\input{sec/0_abstract}    
\input{sec/1_intro}
\input{sec/2_preliminaries}

\input{sec/3_methods}
\input{sec/4_experiments}
\input{sec/5_relation}
\input{sec/6_conclusion}
\input{sec/7_ethics}

{
    \small
    \bibliographystyle{ieeenat_fullname}
    \bibliography{main}
}

\input{sec/X_suppl}

\end{document}

%% file: preamble.tex
\usepackage{times}  
\usepackage{helvet}  
\usepackage{courier}  
\usepackage[hyphens]{url}  
\usepackage{graphicx} 
\usepackage{natbib}  
\usepackage{caption} 
\usepackage{algorithm}
\usepackage{algorithmic}
\usepackage{xcolor}
\usepackage{amsmath}
\usepackage{amssymb}
\usepackage{booktabs}
\usepackage{tabularx}
\usepackage{wasysym}
\usepackage{multirow}
\usepackage{colortbl}

\definecolor{skyblue}{RGB}{0, 176, 240}
\definecolor{orange}{RGB}{234, 99, 51}
\definecolor{neon}{RGB}{128, 245, 16}
\definecolor{violet}{RGB}{102, 43, 148} 
\definecolor{greeen}{RGB}{112, 173, 71}
\definecolor{bluee}{RGB}{83, 112, 255}

\usepackage{pifont}
\newcommand{\cmark}{\ding{51}}%
\newcommand{\xmark}{\ding{55}}%

%% file: sec/0_abstract.tex
\begin{abstract}
Conventional open-world object detection (OWOD) problem setting first distinguishes known and unknown classes and then later incrementally learns the unknown objects when introduced with labels in the subsequent tasks. However, the current OWOD formulation heavily relies on the external human oracle for knowledge input during the incremental learning stages. Such reliance on run-time makes this formulation less realistic in a real-world deployment. To address this, we introduce a more realistic formulation, named semi-supervised open-world detection (SS-OWOD), that reduces the annotation cost by casting the incremental learning stages of OWOD in a semi-supervised manner. We demonstrate that the performance of the state-of-the-art OWOD detector dramatically deteriorates in the proposed SS-OWOD setting. Therefore, we introduce a novel SS-OWOD detector, named SS-OWFormer, that utilizes a feature-alignment scheme to better align the object query representations between the original and  augmented images to leverage the large unlabeled and few labeled data. We further introduce a pseudo-labeling scheme for unknown detection that exploits the inherent capability of decoder object queries to capture object-specific information. On the COCO dataset, our SS-OWFormer using only 50\% of the labeled data achieves detection performance that is on par with the state-of-the-art (SOTA) OWOD detector using all the 100\% of labeled data. Further, our SS-OWFormer achieves an absolute gain of 4.8\% in unknown recall over the SOTA OWOD detector. Lastly, we demonstrate the effectiveness of our SS-OWOD problem setting and approach for remote sensing object detection, proposing carefully curated splits and baseline performance evaluations. Our experiments on 4 datasets including MS COCO, PASCAL, Objects365 and DOTA demonstrate the effectiveness of our approach. Our source code, models and splits are available here {\small\url{https://github.com/sahalshajim/SS-OWFormer}}.
\end{abstract}

%% file: sec/1_intro.tex
\section{Introduction}
\begin{figure}
    \centering
    \includegraphics[width=0.47\textwidth]{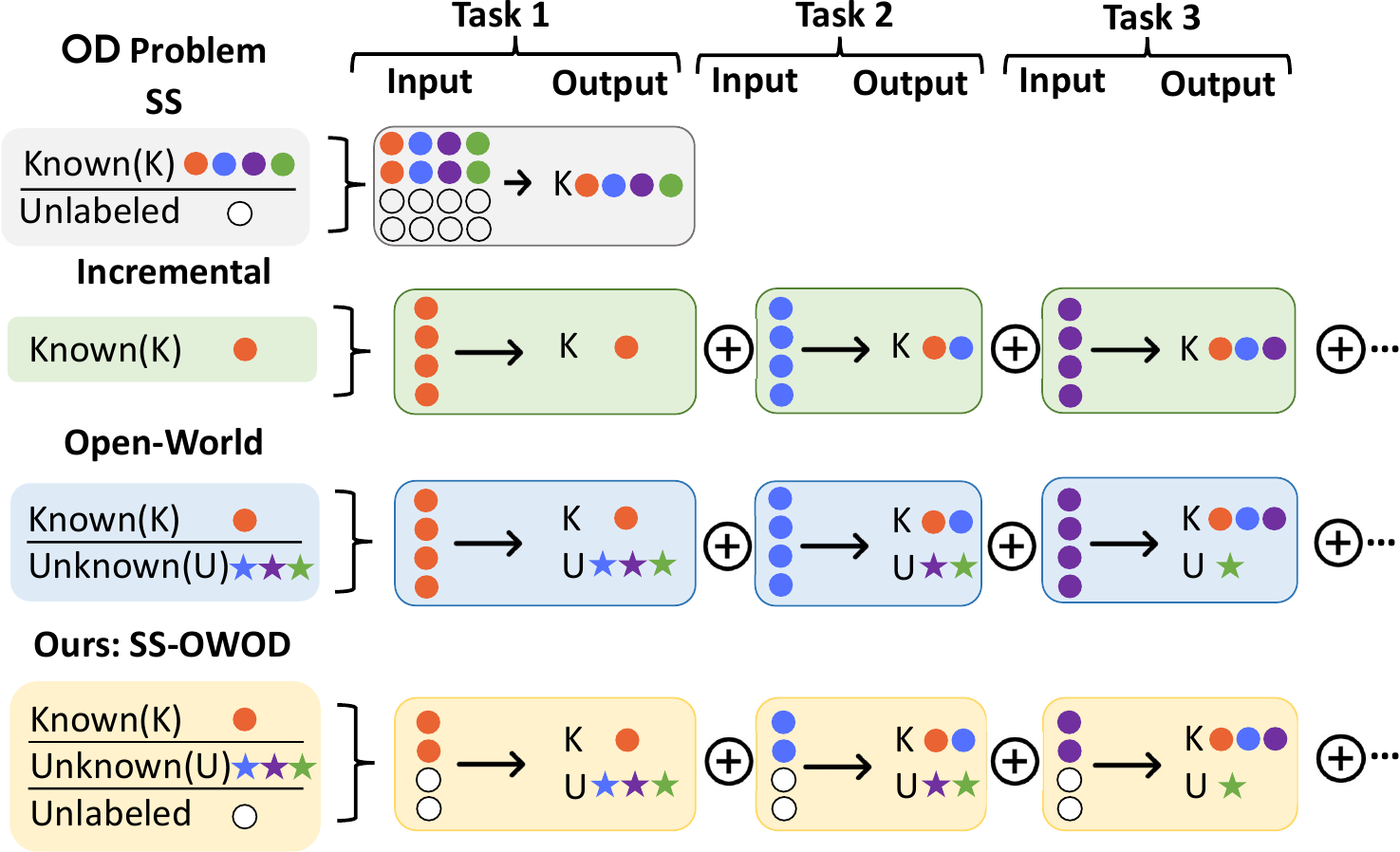}
    \caption{Comparison of our SS-OWOD with other closely related object detection problem settings.}
    \label{fig:problem_figure}
\end{figure}

Conventional object detectors are built upon the assumption that the model will only encounter `known' object classes that it has come across while training \cite{girshick2014rich,carion2020endtoend,zong2023detrs}. Recently, the problem of open-world object detection (OWOD) has received attention \cite{joseph2021open,gupta2021ow}, where the objective is to detect known and `unknown' objects and then incrementally learn these `unknown' objects when introduced with labels in the subsequent tasks. 
In this problem setting, the newly identified unknowns are first forwarded to a human oracle, which can label new classes of interest from the set of unknowns. The model then continues to learn and update its understanding with the new classes without retraining on the previously known data from scratch. Thus, the model is desired to identify and subsequently learn new classes of objects in an \mbox{incremental way when new data arrives}. 

As shown in Fig.\ref{fig:problem_figure} \textit{Semi-supervised} (SS) object detection  learns a set of known classes (\textcolor{orange}{$\bullet$}\textcolor{blue}{$\bullet$}\textcolor{violet}{$\bullet$}\textcolor{greeen}{$\bullet$}), while being fed labeled and unlabeled ($\Circle$) data. In \textit{incremental learning}, classes are learned in steps, as illustrated, the model learns \textcolor{orange}{$\bullet$} in task 1, then fed (\textcolor{blue}{$\bullet$}) in the next task and learns to detect (\textcolor{blue}{$\bullet$}) without forgetting previously learned class (\textcolor{orange}{$\bullet$}), repeating the same process for the subsequent tasks.  \textit{Open-world} object detection aims at detecting unknowns (\textcolor{blue}{$\star$}$\textcolor{violet}{\star}\textcolor{greeen}{\star}$) along with known classes (\textcolor{orange}{$\bullet$}). Unknown classes labeled by a human oracle are learned by the model in the next task as illustrated: the unknown (\textcolor{blue}{$\star$}) is learned as a known (\textcolor{blue}{$\bullet$}) in the next task while continuing to detect remaining unknowns (\textcolor{violet}{$\star$},\textcolor{greeen}{$\star$}). The same procedure is repeated in the subsequent tasks, where the unknown (\textcolor{violet}{$\star$}) is learned as a known (\textcolor{violet}{$\bullet$}). In contrast, we propose the \textbf{SS-OWOD}  setting that aims to reduce the labeling cost of the incoming data of detected unknowns ($\textcolor{blue}{\star}\textcolor{violet}{\star}\textcolor{greeen}{\star}$),by leveraging the unlabeled data ($\Circle$).

Open-world object detection (OWOD) provides a more realistic setting in two ways: (i) It assumes that not all the data in terms of semantic concepts are available during the model training and (ii) it assumes that the data points are non-stationary.
Although standard OWOD provides flexibility to detect unknown object categories and then incrementally learn new object categories, the general problem of incremental learning of new classes comes with the need to be trained in a  \textit{fully supervised} setting \cite{fini2021self}.
To this end, current OWOD approaches rely on strong oracle support to consistently label \textit{all} the identified unknowns with their respective semantics classes and precise box locations.

The objective of this paper is to decrease the aforementioned reliance on the human oracle to provide annotations at run time for the unknown classes (see Fig.\ref{fig:problem_figure}). 
We argue that it is less realistic to assume that an interacting oracle is going to provide annotations for a large amount of data. 
The annotation problem becomes extremely laborious in domains like satellite object detection requiring a much higher number of dense oriented box annotations, in the presence of background clutter and small object size. 
Moreover, existing OWOD methods rely on naive heuristics such as simple  averaging across backbone feature channels \cite{gupta2021ow}  or  clustering of latent feature vectors \cite{joseph2021open} to pseudo-label unknown objects, thereby struggling to accurately detect the unknowns.  
To this end, we propose a novel transformer-based method, named SS-OWFormer,  that collectively addresses both the issues of improving unknown detection and reducing the annotation cost for identified unknowns during the life span of model learning. 

\noindent\textbf{Contributions:} 
The primary contributions of this research encompass the following aspects: \\
\textbf{(i)} We introduce a novel Semi-supervised  Open-World Object Detection (SS-OWOD) problem setting that reduces the strong dependence on external human oracles to provide annotations for \textit{all} incoming data in incremental learning stages. We further propose a \textit{Semi-supervised Open-World object detection Transformer} framework, named, SS-OWFormer, designed to detect a newly introduced set of classes in a semi-supervised open-world setting. 
SS-OWFormer utilizes a feature alignment scheme to effectively align the object query representations between the original and augmented copy of the image for exploiting the large unlabeled and fewer labeled data. \\
\textbf{(ii)} We introduce a pseudo-labeling scheme to better distinguish the unknown objects by exploiting the inherent capability of the detection detector object queries to capture object-specific information. The resulting modulated object queries provide multi-scale spatial maps to obtain the objectness confidence scores which in turn are used for \mbox{the pseudo-labeling process.} \\
\textbf{(iii)} Comprehensive experiments on OWOD COCO split \cite{joseph2021open} are performed to demonstrate the effectiveness of our approach. Our SS-OWFormer achieves favorable detection performance for both the `known' and `unknown' classes in all the tasks, compared to the state-of-the-art OW-DETR~\cite{gupta2021ow}.  SS-OWFormer achieves superior overall detection performance when using only 10\% of the labeled data, over OW-DETR using 50\% labeled data. In terms of `unknown' detection, SS-OWFormer achieves an absolute gain of $4.8\%$, in terms of \mbox{unknown recall, over OW-DETR.} \\
\textbf{(iv)} Lastly, we explore for the first time the SS-OWOD problem for remote sensing domain. We show the effectiveness of our SS-OWFormer on satellite images, where the labeling task is even more laborious and time-consuming. 
Moreover, we have proposed open world splits for the Object365 dataset having large number of categories. Our experiments on 4 datasets including MS COCO, PASCAL, Objects365 and DOTA demonstrate the \mbox{effectiveness of our approach.}

 \begin{figure}
    \centering
    \includegraphics[width=0.47\textwidth]{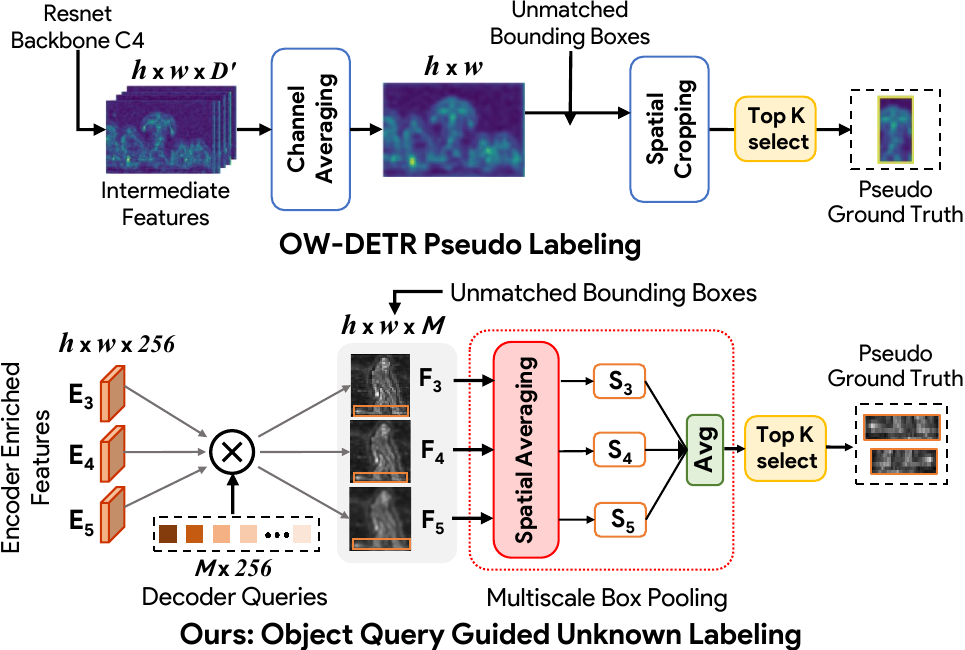}
    \caption{Comparison of our object query guided pseudo-labeling with feature averaging used in OW-DETR baseline. The baseline framework performs a channel averaging over single-scale features from the backbone, spatially crops them at the predicted bounding box positions, and selects the top-k to obtain pseudo-labels. In contrast, our approach strives to leverage object-specific information from multi-scale encoder features \textit{and} decoder object queries.  We modulate the decoder object queries with multi-scale encoder feature maps and perform multi-scale box pooling at predicted box locations to obtain objectness scores and select the top-k bounding box proposals as pseudo labels.}
    \label{fig:pseudo_labeling}
\end{figure}

%% file: sec/2_preliminaries.tex
\section{Preliminaries}
Let $\mathcal{{D}}^t = \{\mathcal{I}^t, \mathcal{Y}^t\}$ be a dataset containing $N$ images $\mathcal{I}^t=\{I_1, I_2,  ..., I_{N}\}$ with corresponding labels ${\mathcal{Y}}^t = \{Y_1, Y_2, ..., Y_{N}\}$. Here, each image label $Y_i = \{y_1, y_2, ..., y_k\}$ is a set of  box annotations for all $k$ object instances in the image.
The open-world object detection (OWOD) follows the incremental training stages on the progressive dataset $\mathcal{{D}}^t$ having only  $\mathcal{K}^{t} = \{{C_{1}, C_{2}, ..., C_{n}}\}$ known object classes at time $t$. A model trained on these $\mathcal{K}^{t}$ known classes is expected to not only detect known classes but also detect (localize and classify) objects from unknown classes $\mathcal{U} = \{C_{n+1},...\}$ by predicting an unknown class label for all unknown class instances. An overview of closely related object detection \mbox{settings is shown in Fig.\ref{fig:problem_figure}}.\\
\noindent\textbf{Proposed SS-OWOD Problem Setting:}
Here, each image label $Y_i = \{y_1, y_2, ..., y_k\}$ is a set of box annotations for all $k$ object instances in the image. The instance annotation $y_k = [l_k, o^{x}_k, o^{y}_k, h_k, w_k]$ consists of $l_k \in \mathcal{K}^t$ is the class label for a bounding box having a center at ($o^{x}_k, o^{y}_k$), width $w_k$, height $h_k$.
In this work, we argue that it is laborious and time-consuming for the human oracle to obtain bounding box annotations for all training images used for learning.
Hence,  we propose a new \textit{semi-supervised open-world object detection} problem setting, where only a partial set of images  ($N_s$) are annotated by the human oracle and the remaining $N_u$ images are unlabeled (see Fig.\ref{fig:problem_figure}).
This aims to reduce the strong dependence on the human oracle for adding knowledge to the model's learning framework.
Here, during learning stages in an open-world setting, the model is expected to utilize both labeled and unlabeled sets of training images ($N_s$+$N_u$) to learn about the new $\mathcal{K}^{t+1}$  classes, without forgetting previously known $\mathcal{K}^{t}$  classes, thereby enabling detection of unknown objects at the same time. 
\subsection{Baseline Framework}
We base our approach on the recently introduced OW-DETR
 \cite{gupta2021ow}. 
It comprises a backbone network, transformer encoder-decoder architecture employing deformable attention, box prediction heads, objectness, and novelty classification branches to distinguish unknown objects from known and background regions.  
Here, the transformer decoder takes a set of learnable object queries as input and employs interleaved cross- and self-attention modules to obtain a set of object query embeddings. 
These object query embeddings are used by the prediction head for box predictions as in \cite{zhu2020deformable}.   It selects bounding boxes of potential unknown objects through a  pseudo-labeling scheme and learns a classifier to categorize these potential unknown object query embeddings into a single unknown class as in \cite{gupta2021ow}. 
 Here, potential unknown objects are identified based on average activations at a selected layer (C4 of ResNet50) of the backbone feature map at regions corresponding to predicted box locations (see Fig.\ref{fig:pseudo_labeling}).  
 Among all potential unknown object boxes, only boxes that are non-overlapping with the known ground-truth boxes are considered pseudo-labels for potential unknowns.  
 It learns a binary class-agnostic objectness branch to distinguish object query embeddings of known and potential unknown objects from background regions. 
 In addition, it learns a novelty classification branch having unknown as an additional class along with $\mathcal{K}^{t}$ known classes as in \cite{gupta2021ow}. 
 We refer to this as our baseline framework.
 
\noindent\textbf{Limitations:}
 As discussed above, the baseline framework employs a heuristic method for pseudo-labeling with a simple averaging across channels of a single-scale feature map to compute objectness confidence where only single-scale features from the backbone are utilized. 
 However, such a feature averaging to identify the presence of an object at that spatial position is sub-optimal for the accurate detection of unknown objects. To improve unknown object detection, it is desired to leverage the object-specific information available in both deformable encoder and decoder features (see Fig.\ref{fig:pseudo_labeling}). Existing state-of-the-art OWOD frameworks, including our baseline, typically require bounding box supervision for \textit{all} images used during the incremental learning of novel classes in the OWOD tasks.  However, this makes the OWOD model strongly dependent on an external human oracle to provide dense annotations for all the data in the subsequent tasks.  Next, we introduce our SS-OWFormer approach that collectively addresses the above issues in a single framework.

%% file: sec/3_methods.tex
\section{Method}

\begin{figure*}[!ht]
\centering
    \includegraphics[scale = 0.70]{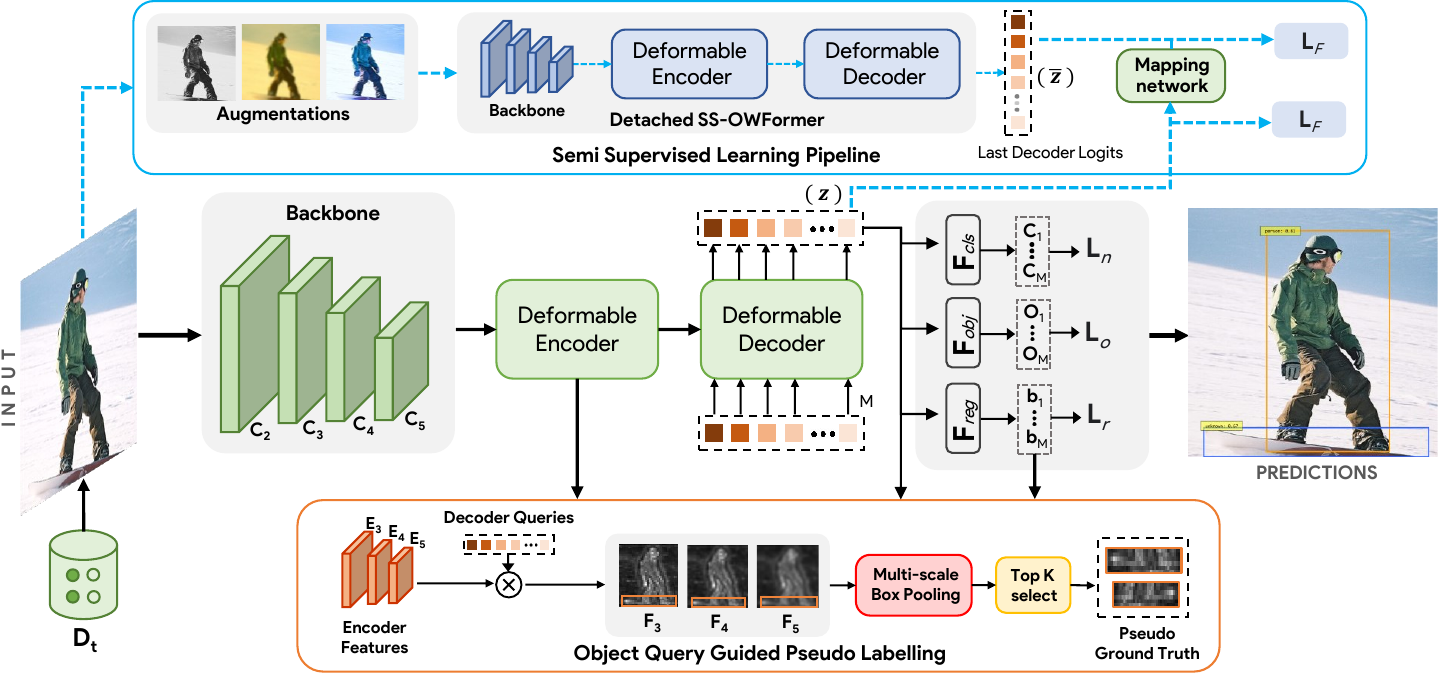}   
    \caption{Overall architecture of our Semi-Supervised Open-World object detection Transformer (SS-OWFormer) framework. It comprises a backbone network, transformer-based deformable encoder-decoder, object query-guided pseudo-labeling, box prediction head, novelty classification, and objectness branches. The focus of our design is: (i) the introduction of a \textit{object query-guided pseudo-labeling} (\textit{orange} box at bottom row) that captures information from both transformer encoder and decoder for pseudo-labeling unknown objects. Object queries from the decoder are modulated with the multi-scale encoder features to obtain multi-scale spatial maps which are pooled at predicted box locations to obtain confidence scores for the unknown pseudo-labeling. (ii) The introduction of a novel semi-supervised learning pipeline   ($\rightarrow$) for leveraging unlabelled data during incremental learning of a new set of object classes. In our  \textit{semi-supervised incremental learning} setting,  the SS-OWFormer (current model) is trained along with its detached (frozen) copy (\textit{blue} box on top row) together with a mapping network ($\mathcal{G}$).  The mapping network ($\mathcal{G}$) projects the object queries from the current network to the detached network.   Moreover, we use original and augmented images for the alignment of object query embeddings ($z$).} 
    \label{fig: architecture}
\end{figure*}

\subsection{Overall Architecture}
\label{ssec:overall_arch}
Fig.\ref{fig: architecture} shows the overall architecture of our \textit{Semi-supervised  Open-World object detection Transformer (SS-OWFormer)} framework. It comprises a backbone network, deformable encoder, deformable decoder, object query-guided pseudo-labeling, and prediction heads. 

The backbone takes an input image of spatial resolution $H \times W$ and extracts multi-scale features for the deformable encoder-decoder network having  $M$ learnable object queries at the decoder. The decoder employs interleaved cross- and self-attention and outputs $M$ object query embeddings ($z$). These query embeddings are used in the box prediction head, objectness and novelty classification branches. In addition, these query embeddings  ($z$) are used in our semi-supervised learning framework to align the current network  (${\mathcal{M}}_{cur}/z$) with a detached network  from the previous task ($\bar{\mathcal{M}}_{prev}/\bar{z}$).  We take augmented images as input to the current network and corresponding query embeddings ($z^a$) are transformed to the latent space of the detached network using a mapping network ($\mathcal{G}$). These transformed embeddings are aligned with the embeddings  ($\bar{z}$) obtained for the same images from the detached network using a feature-aligning strategy detailed in Sec.\ref{sub:ssil}. 

We employ fully supervised learning for the first task (task-1) where the object detector is trained with initial known object categories. During task-1 inference, the model is expected to  detect all known and unknown object categories. Then, in the subsequent task, the model is trained with new  object categories in our novel semi-supervised incremental learning setting  where we have annotations only for a partial set of training data. Here, the objective is to learn new object categories using  labeled and unlabeled data  without forgetting the task-1 categories.  To this end, we use a detached network  whose weights are fixed during our incremental learning and an identical current network where the network weights are updated. We learn the current network (by taking the detached network as a reference)  using labeled and unlabeled data, followed by fine-tuning the current network using available labeled data. Next, we introduce our object query-guided \mbox{unknown-labeling scheme.}

\subsection{Object Query Guided Pseudo-Labeling}
As discussed, we need to accurately detect unknown objects out of the known set of classes in open-world object detection. Here, the model is expected to transfer its known object detection knowledge to detect unknown objects. Our baseline utilizes a single-scale pseudo-labeling scheme which is a simple heuristic approach with a naive way of averaging Resnet features for pseudo-labeling unknowns. We aim to utilize learnable properties intrinsic to the deformable transformer architecture from encoder features and decoder queries. This is found to be more suitable for the objectness confidence levels to be used for pseudo-labeling. 
Let ${F} = \{\text{E3}, \text{E4}, \text{E5} \}$ be multi-scale encoder features and  $y_k = [o^{x}_k,o^{y}_k,h_k,w_k]$ be a box proposal predicted for a given object query embedding.  
Let $\text{E}_i\in R^{H_i\times W_i\times D}$ be the encoder feature map at scale i and $M$ queries $Q_j\in R^{M\times D}$  be the unmatched object queries at the decoder. 
Then, we  modulate the  encoder features with a transposed matrix multiplication to obtain query-modulated feature maps $\mathcal{F}_i\in R^{H_i\times W_i\times M}$. 
This query-modulated feature map results in better scoring for objectness since it leverages object-specific information from decoder queries along with encoder features. 
Then, we perform multi-scale box pooling over these maps $\mathcal{F}_i$ at predicted box locations of respective object queries. Our  multi-scale box pooling  performs spatial averaging over these  spatial maps $\mathcal{F}_i$ to obtain objectness scores $s_k$ corresponding to bounding boxes. For instance, the objectness score for a bounding box (b) can be calculated as,
\begin{equation}
\small
\begin{split}
    \sum_{i=0}^{n} S_i (b) & = \frac{1}{h_b \cdot w_b} \sum_{i=0}^{n} \mathcal{F}_i\\
                       & = \frac{1}{h_b \cdot w_b} \sum_{i=0}^{n} E_i \cdot \sum_{j=0}^{M-K}Q_j^{T}  
\end{split}
\end{equation}

These objectness scores are used to select the top $k$ boxes which are then used as pseudo-labels to train the novelty classifier and objectness branches.  
The regression branch in the prediction head takes $M$ object query embeddings from the decoder and predicts $M$  box proposals.  The bipartite matching loss in the decoder selects $K$ queries (from $M$ total queries) as positive matches for the known classes \mbox{in the supervised setting.} 

\subsection{Semi-supervised Open-world Learning} 
\label{sub:ssil}
Previous open-world object detection works assume that all incoming data for novel classes are labeled while in a realistic scenario, it might prove to be costly. 
However, in our semi-supervised open-world object detection formulation, we employ semi-supervised learning for incremental learning. So, in our challenging setting, the model has to learn to detect novel object categories  by using a limited amount of partially annotated  data along with unlabeled data for the novel classes and detecting unknown objects, without forgetting  previously learned categories. 

As discussed in Sec.\ref{ssec:overall_arch}, we introduce a subset of object categories to the model through subsequent tasks.   For the first task, the model is trained like a standard OW object detector, and a set of classes $\mathcal{K}^1 = \{C_1, C_2, ... C_{n}\}$ are introduced.  
Then for the subsequent tasks, semi-supervised learning is leveraged for the limited availability of annotations. Using a detached copy of the model from the previous task, $\bar{\mathcal{M}}_{pre}$, the current model ${\mathcal{M}}_{cur}$ is trained on labeled and unlabeled data with a \textit{feature-aligning strategy}. 

For semi-supervised learning using the next progressive dataset $D^{t+1}$, we employ strong augmentations such as color-jitter, random greyscaling, and blurring to obtain augmented data $\mathcal{D}_a^{t+1} = \{\mathcal{I}^{a} \}$. The augmentations here are selected such that they do not change the  box positions in input images, hence better suitable for semi-supervised object detection.  Furthermore,  we do not use augmentations such as rotation, flipping, translation, cropping, etc that are likely to alter the feature representation in augmented images. 
We use a \textit{detached model} $\bar{\mathcal{M}}_{pre}$ whose weights are fixed, a current model ${\mathcal{M}}_{cur}$ with learnable weights, and a  mapping network $\mathcal{G}$ that maps the current model object queries to the detached model object queries.   Here, a copy of the current model with fixed weights is used as the \textit{detached model} $\bar{\mathcal{M}}_{pre}$. This detached model does not receive any gradient and remains detached during training. 

For an image $I_i$ from  $D^{t+1}$, and its augmented version $I^a_i$ from  $\mathcal{D}_a^{t+1}$,  we extract object query  features using the current and detached models. i.e, We use the current model and obtain original image object query embedding feature  $z = {\mathcal{M}}_{cur}(I_i)$ and  augmented image query embedding, $z^a = {\mathcal{M}}_{cur}(I_i^a)$. 
Similarly,   the detached model is used to obtain the embedding $\Bar{z}^a = \bar{\mathcal{M}}_{pre}(I_i^a)$.
Then, our mapping network $\mathcal{G}$ maps $z^a$  to $\Bar{z}^a$ 
instead of enforcing $z^a$ to be similar to $\Bar{z}^a$ as that may adversely affect the learning in the distillation loss $\mathcal{L}_D$.  We perform  feature alignment to bring the object queries  $\mathcal{G}\left(z^a\right)$ and  $\bar{z}^a$ together by employing a feature alignment loss $\mathcal{L}_{F}$.
Here, we measure the cross-correlation matrix ~\cite{zbontar2021barlow} between input embeddings and try to bring the object queries  closer. The loss also helps to reduce the redundancy between embeddings and makes the representations robust to noise. 
In addition, the same loss is used to make the model invariant to augmentation, which in turn may help the object query representations $z$ invariant to the state of the model. Then, the current model  ${\mathcal{M}}_{cur}$ is trained using the following loss:
\begin{equation}
\small
    \begin{aligned}
\mathcal{L}_{cur} &=\mathcal{L}_{F}\left(\boldsymbol{z}^a, \boldsymbol{z}\right)+\mathcal{L}_D\left(\boldsymbol{z}^a, \boldsymbol{\bar{z}^a}\right) \\
&=\mathcal{L}_{F}\left(\boldsymbol{z}^a, \boldsymbol{z}\right)+\mathcal{L}_{F}\left(\mathcal{G}\left(\boldsymbol{z}^a\right), \boldsymbol{\bar{z}^a}\right)
\end{aligned}
\label{eq:bt}
\end{equation}

 \begin{figure}[t!]
    \centering
    \includegraphics[scale = 0.57]{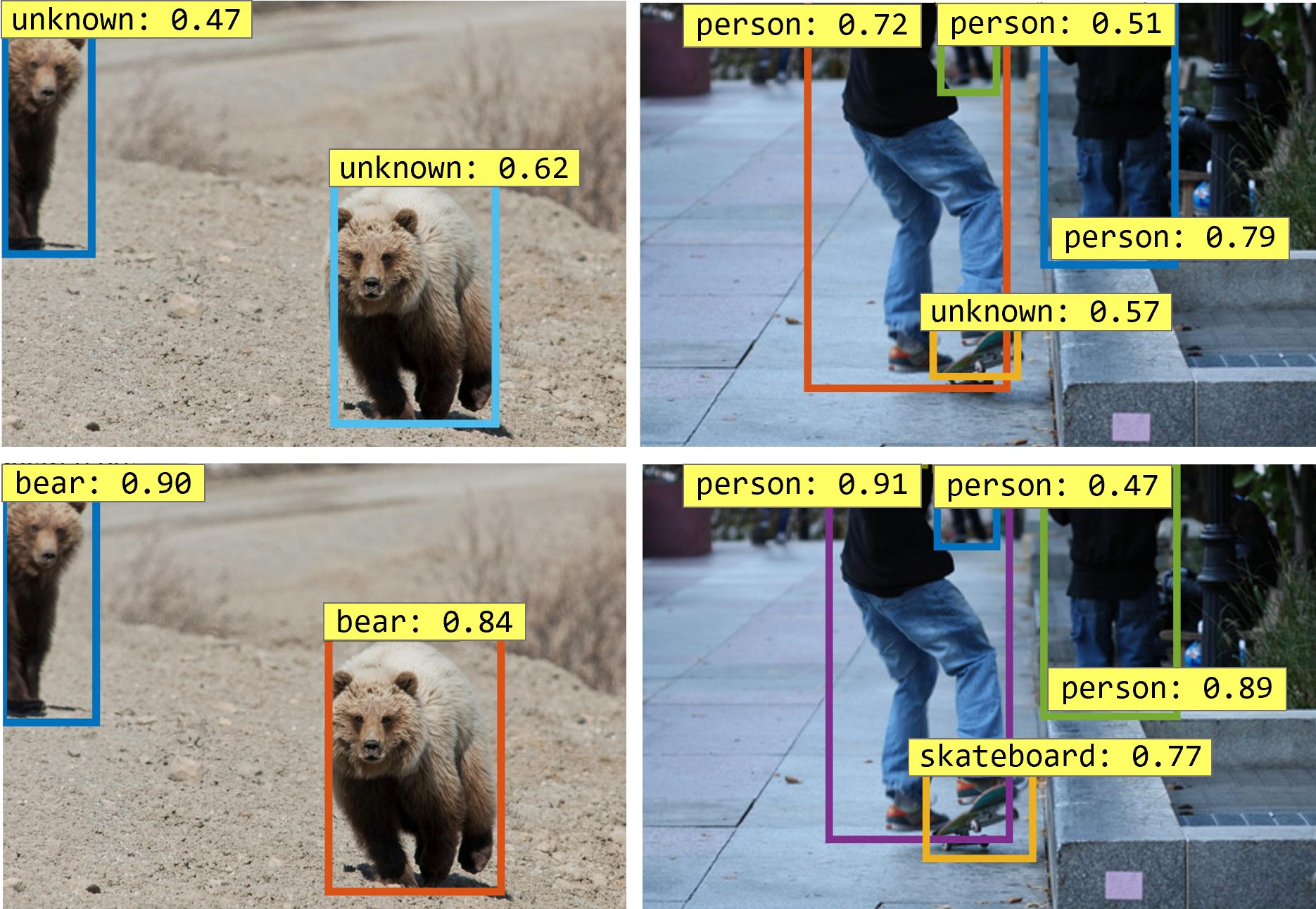} 
    \caption{Qualitative results showing the detection performance on MS COCO examples. 
    From the top row, the unknown classes are learned to be marked as a known category in the subsequent tasks as shown in the bottom row.}
    \label{fig:coco_viz}
\hfill
\end{figure} 

\begin{figure}
    \centering
    \includegraphics[width=0.47\textwidth]{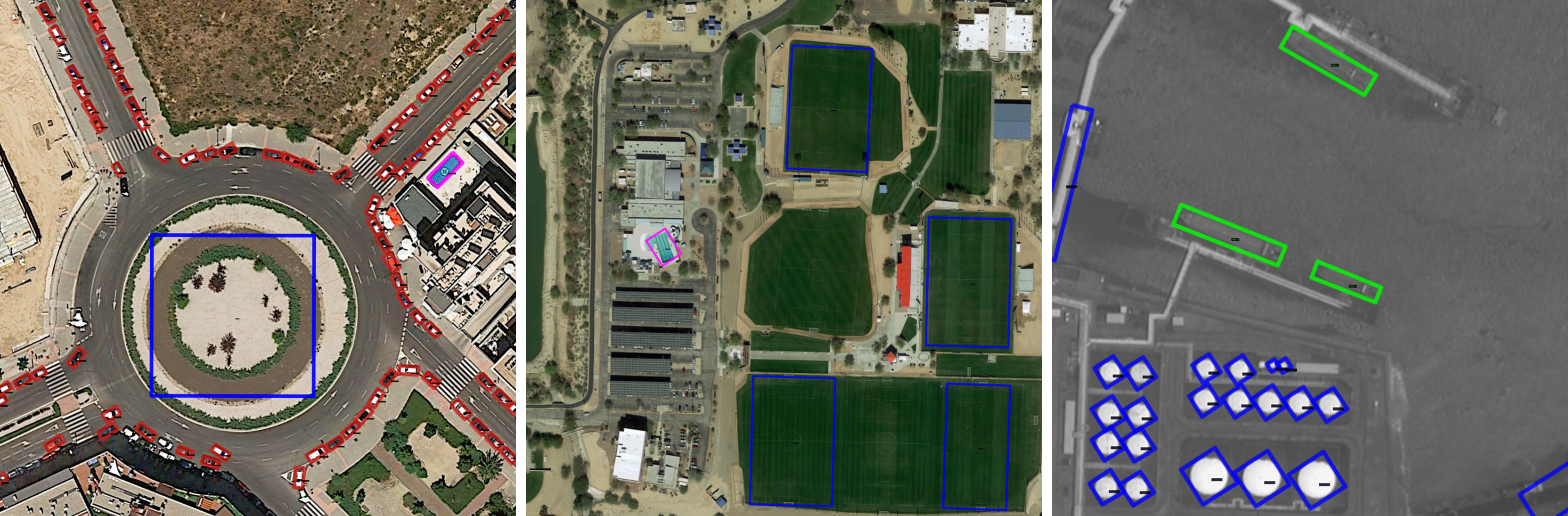}
    \caption{Qualitative results on satellite images with oriented bounding boxes. Oriented bounding boxes in \textit{blue} depict unknown detections on the categories of roundabout, soccer field, and storage tanks in the images respectively. While other colors mark known categories of small-vehicle, swimming pool, and ship.}
    \label{fig:sat_viz}
\end{figure}

\subsection{OWOD in Satellite Images}
Different from the OW-DETR that predicts axis-parallel bounding boxes in natural images, for satellite images we adapt our baseline framework to predict oriented bounding boxes along object directions for a more generalizable approach. 
For oriented object detection, we introduce an additional angle prediction head in OW-DETR and its standard bounding box prediction heads.
Our Object Query Guided Pseudo-Labeling scheme is also found to be suitable for the challenges presented in satellite imagery such as large-scale variations, high object density, heavy background clutter,  and a large number of  object instances in satellite images.
Moreover, the dependence on human oracles for open-world object detection in satellite imagery is highly problematic because of the requirement of a high number of dense-oriented bounding box annotations per image. Thereby, a semi-supervised open-world learning \mbox{setting can prove beneficial.}

\begin{table*}[t!]
\centering
\begin{tabular}{|l|cccc|cccc|ccc|}
\hline
\multirow{3}{*}{\textbf{Method}} & \multicolumn{4}{c|}{\textbf{Task2}}                                                                                                                  & \multicolumn{4}{c|}{\textbf{Task3}}                                                                                                                  & \multicolumn{3}{c|}{\textbf{Task4}}                                                        \\ \cline{2-12} 
                        & \multicolumn{1}{c|}{\multirow{2}{*}{\small\textbf{U-Recall}}} & \multicolumn{3}{c|}{\small\textbf{mAP}}                                                          & \multicolumn{1}{c|}{\multirow{2}{*}{\small\textbf{U-Recall}}} & \multicolumn{3}{c|}{\small\textbf{mAP}}                                                          & \multicolumn{3}{c|}{\small\textbf{mAP}}                                                          \\
                        & \multicolumn{1}{c|}{}                                   & \multicolumn{1}{c|}{\small\textbf{Prev}}  & \multicolumn{1}{c|}{\small\textbf{Cur}}   & \small\textbf{Both}  & \multicolumn{1}{c|}{}                                   & \multicolumn{1}{c|}{\small\textbf{Prev}}  & \multicolumn{1}{c|}{\small\textbf{Cur}}   & \small\textbf{Both}  & \multicolumn{1}{c|}{\small\textbf{Prev}}  & \multicolumn{1}{c|}{\small\textbf{Cur}}   & \small\textbf{Both}  \\ \hline \hline
\small\textbf{ORE-EBUI}       & \multicolumn{1}{c|}{2.9}                                & \multicolumn{1}{c|}{52.7}           & \multicolumn{1}{c|}{26}             & 39.4           & \multicolumn{1}{c|}{3.9}                                & \multicolumn{1}{c|}{38.2}           & \multicolumn{1}{c|}{12.7}           & 29.7           & \multicolumn{1}{c|}{29.6}           & \multicolumn{1}{c|}{12.4}           & 25.3           \\ 
\small\textbf{OW-DETR}        & \multicolumn{1}{c|}{6.2}                                & \multicolumn{1}{c|}{53.6}           & \multicolumn{1}{c|}{33.5}           & 42.9           & \multicolumn{1}{c|}{5.7}                                & \multicolumn{1}{c|}{38.3}           & \multicolumn{1}{c|}{15.8}           & 30.8           & \multicolumn{1}{c|}{31.4}           & \multicolumn{1}{c|}{17.1}           & 27.8           \\ \hline
\small\textbf{OW-DETR (50\%)} & \multicolumn{1}{c|}{6.94}                               & \multicolumn{1}{c|}{50.53}          & \multicolumn{1}{c|}{19.28}          & 34.91          & \multicolumn{1}{c|}{7.64}                               & \multicolumn{1}{c|}{32.7}           & \multicolumn{1}{c|}{9.13}           & 24.85          & \multicolumn{1}{c|}{24.08}          & \multicolumn{1}{c|}{5.74}           & 19.49          \\ 
\small\textbf{SS-OWFormer (50\%)}             & \multicolumn{1}{c|}{\textbf{10.56}}                     & \multicolumn{1}{c|}{\textbf{52.04}} & \multicolumn{1}{c|}{\textbf{26.35}} & \textbf{39.2}  & \multicolumn{1}{c|}{\textbf{13.16}}                     & \multicolumn{1}{c|}{\textbf{39.46}} & \multicolumn{1}{c|}{\textbf{13.63}} & \small\textbf{30.85} & \multicolumn{1}{c|}{\textbf{29.97}} & \multicolumn{1}{c|}{\textbf{11.48}} & \textbf{25.35} \\ \hline
\small\textbf{OW-DETR (25\%)} & \multicolumn{1}{c|}{5.03}                               & \multicolumn{1}{c|}{49.19}          & \multicolumn{1}{c|}{15.64}          & 32.42          & \multicolumn{1}{c|}{6.94}                               & \multicolumn{1}{c|}{31.02}          & \multicolumn{1}{c|}{9.13}           & 23.72          & \multicolumn{1}{c|}{22.9}           & \multicolumn{1}{c|}{6.39}           & 18.77          \\ 
\small\textbf{SS-OWFormer (25\%)}    & \multicolumn{1}{c|}{10.47}                              & \multicolumn{1}{c|}{52.21}          & \multicolumn{1}{c|}{21.16}          & 36.68          & \multicolumn{1}{c|}{12.22}                              & \multicolumn{1}{c|}{36.4}           & \multicolumn{1}{c|}{10.83}          & 27.87          & \multicolumn{1}{c|}{26.91}          & \multicolumn{1}{c|}{8.72}           & 22.36          \\ \hline
\small\textbf{OW-DETR (10\%)} & \multicolumn{1}{c|}{4.83}                               & \multicolumn{1}{c|}{47.8}           & \multicolumn{1}{c|}{12.36}          & 30.08          & \multicolumn{1}{c|}{8.24}                               & \multicolumn{1}{c|}{30.65}          & \multicolumn{1}{c|}{6.14}           & 22.48          & \multicolumn{1}{c|}{21.23}          & \multicolumn{1}{c|}{4.78}           & 17.11          \\ 
\small\textbf{SS-OWFormer (10\%)}            & \multicolumn{1}{c|}{\textbf{10.19}}                      & \multicolumn{1}{c|}{\textbf{53.61}} & \multicolumn{1}{c|}{\textbf{16.44}} & \textbf{35.02} & \multicolumn{1}{c|}{\textbf{12.13}}                     & \multicolumn{1}{c|}{\textbf{35.21}} & \multicolumn{1}{c|}{\textbf{8.11}}  & \textbf{26.18} & \multicolumn{1}{c|}{\textbf{26.17}} & \multicolumn{1}{c|}{\textbf{5.33}}  & \textbf{20.96} \\ \hline
\end{tabular}   
\caption{State-of-the-art comparison for the open-world object detection (OWOD) problem on natural images using MS COCO split of \cite{joseph2021open}. The comparison is presented in terms of unknown recall (U-Recall) and the previously known (Prev), current known (Cur) and Overall (both) AP for all tasks. U-Recall is not reported for task-4 since all classes are known. 
Our SS-OWFormer with just $10\%$ labeled data outperforms the SOTA OW-DETR with $50\%$ labeled data on all tasks.} 
\label{tab:sota}
\end{table*}

\subsection{Training and Inference}
\noindent \textbf{Training:} 
The overall loss formulation for the network can be written as:
\begin{equation}
\small
    \mathcal{L} = \mathcal{L}_c + \mathcal{L}_r + \alpha\mathcal{L}_o + \mathcal{L}_{cur}
\end{equation}
where $\mathcal{L}_c$, $\mathcal{L}_r$ and $\alpha\mathcal{L}_o$ respectively denote classification, bounding box regression, foreground objectness (class-agnostic) loss terms while $\mathcal{L}_{cur}$ stands for the loss from semi-supervised incremental learning from eq. \ref{eq:bt}.

The proposed framework follows multi-stage training. The first task is trained in a fully supervised manner using   
$\mathcal{L}_c $, $\mathcal{L}_r$, $\mathcal{L}_o$. 
Then, the subsequent tasks follow the \textit{feature alignment} strategy using an additional $\mathcal{L}_{cur}$ loss.  A detached model and a current model are trained on augmented unannotated data together with a mapping network $\mathcal{G}$ on top to bring the embeddings closer in latent space \mbox{using feature-alignment.}\\
\textbf{Inference:} The object queries for a test image $I$ are obtained and the model predicts their labels from $\mathcal{K}^t + 1$ classes along with a bounding box. A \textit{top-k} selection with the highest scores is used for OWOD detection.

\begin{figure}
    \centering
    \includegraphics[width=0.47\textwidth]{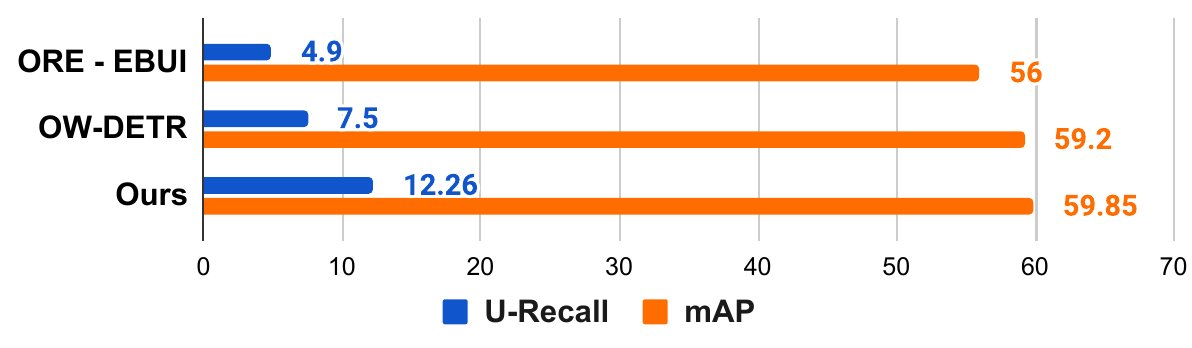}
    \caption{State-of-the-art comparison for OWOD Task-1. U-Recall reports the performance of the model in detecting unknown classes and mAP evaluates the performance of known classes. Owing to object query-guided pseudo-labeling our framework SS-OWFormer outperforms SOTA OW-DETR in terms of \mbox{U-Recall and mAP.}} 
    \label{fig:task1}
\end{figure}

%% file: sec/4_experiments.tex
\section{Experiments}
\noindent\textbf{Datasets:} We evaluate our SS-OWOD framework on MS-COCO \cite{lin2014microsoft}, Pascal VOC \cite{everingham2010pascal}, DOTA \cite{xia2018dota} and Objects365 \cite{shao2019objects365} for OWOD problem. Non-overlapping tasks $\{T_1, T_2, ..., T_t, ...\}$ are formed from classes such that a class in $T_\lambda$ is not introduced till $t=\lambda$ is reached as introduced in \cite{joseph2021open}. 
Our novel satellite OWOD split is prepared from DOTA based on the number of instances per image to 
ensure the corresponding \mbox{representation of all categories.}\\
\noindent\textbf{Evaluation Metrics: }The standard mean average precision (mAP) is the metric for known classes. However, mAP cannot be utilized as a fair metric to evaluate unknown detection since all possible unknowns are not labeled and can be more than given classes to be encountered in the next set of tasks. Therefore, we use average recall as a metric to test unknown object detection, as in \cite{bansal2018zero, lu2016visual} \mbox{under a similar context.}\\
\noindent\textbf{Implementation Details:} The transformer architecture is a version of Deformable DETR \cite{zhu2020deformable}. Multi-scale feature maps are taken from ImageNet pre-trained ResNet50 \cite{he2016deep,zhang2022dino}. Number of queries is set to $M = 250$ to account for the high number of instances in satellite images, while the threshold for the selection of pseudo-labels is set to top-10. Training is carried out for 50 epochs using ADAM optimizer \cite{kingma2014adam} with weight decay (AdamW) and learning \mbox{rate set to $10^{-4}$}.

\begin{table}[t]
\centering
    \begin{tabular}{|l|c|c|}
    \hline
    \textbf{Method}               & \textbf{U-Recall} & \textbf{mAP} \\ \hline \hline
    baseline                       & 6.17              & 58.53        \\
    + Enc\_feature           & 8.06              & 58.56        \\
    SS-OWFormer             & \textbf{12.26 }            & 59.85        \\ \hline
    SS-OWFormer(Max)             &{11.25}            & 59.51        \\ 
    SS-OWFormer(Mean)         & \textbf{12.26 }            & 59.85        \\ \hline
    \end{tabular}
    \caption{Ablation analysis of the impact on performance with various design choices for pseudo-labeling. The bottom section shows design choices for the selection of objections scores.}
    \label{tab:ablation}
\end{table} 

\begin{table}[t]
\centering
\begin{tabular}{|l|c|c|c|}
\hline
\textbf{Model}                                                                    & \textbf{Evaluation} & \textbf{mAP} & \textbf{U-Recall} \\ \hline \hline
\multirow{2}{*}{Baseline}                                                         & Task-1              & 64.9                & 2.5                     \\
                                                                                  & Task-2               & 68.1                & -                       \\ \hline
\multirow{2}{*}{SS-OWFormer} & Task-1              & 66.7                & \textbf{7.6}                     \\
                                                                                  & Task-2              & \textbf{70.9 }               & -                       \\ \hline
\end{tabular} 
\caption{Comparison between the baseline and our SS-OWFormer on the proposed satellite OWOD splits. The comparison is shown in terms of mAP and unknown recall. The unknown recall metric assesses the model's ability to capture unknown object instances. While using full task-2 data, SS-OWFormer improves 5.1\% in unknown recall over baseline without compromising  overall mAP.}\vspace{-0.1cm}
\label{tab:sat1}
\end{table}

\begin{table}[t!]
\centering
\begin{tabular}{|c|c|c|c|c|}
\hline
\multirow{2}{*}{\textbf{\begin{tabular}[c]{@{}c@{}}\\ Partial \\ Annotation\end{tabular}}} & \multirow{2}{*}{\textbf{SSL}} & \multicolumn{3}{c|}{\textbf{mAP}}                                                                                                                       \\ \cline{3-5} 
                                                                                        &                               & \textbf{Overall} & \textbf{\begin{tabular}[c]{@{}c@{}}Previously \\ Known\end{tabular}} & \textbf{\begin{tabular}[c]{@{}c@{}}Current \\ Known\end{tabular}} \\ \hline \hline
100\%                                                                                   & \cmark                           & 70.9             & 77.1                                                              & 61.6                                                            \\ \hline
\multirow{2}{*}{75\%}                                                                   & \xmark                           & 65.2             & 74.8                                                              & 50.9                                                            \\
                                                                                        & \cmark                          & 69.04             & 76.2                                                              & 58.3                                                           \\ \hline
\multirow{2}{*}{50\%}                                                                   & \xmark                           & 63               & 74.5                                                              & 45.8                                                            \\
                                                                                        & \cmark                          & 68.06             & 75.1                                                              & 57.5                                                        \\ \hline

\end{tabular} 
\caption{Comparison of results with and without semi-supervised learning on proposed satellite OWOD splits. This demonstrates  our semi-supervised incremental learning strategy at different proportions of partially annotated data with a steady improvement over baseline under similar settings. The semi-supervised learning pipeline enables us to take advantage of the unannotated data while maintaining the performance on previously known classes. } 
\label{tab:sat2}
\end{table}

\subsection{State-of-the-art Comparison} \label{subsec: exp_nat}
 We show a comparison with previous works \cite{joseph2021open, gupta2021ow} in OWOD for splits on MS COCO in Tab.\ref{tab:sota}. The qualitative results can be seen in Fig.\ref{fig:coco_viz}.
 The comparison is made in terms of known class mAP and unknown class recall (U-Recall). U-Recall quantifies the model's capacity to retrieve unknown object instances in the OWOD setting. It should be noted that, since all classes are known in task-4, U-Recall cannot be reported.
 For a fair comparison, we omit ORE's energy-based unknown identifier (EBUI) since that relies on a held-out validation set.
 Our contributions prove useful in a fully supervised setting for task-1 as depicted above in Fig.\ref{fig:task1}. 
 Compared to the state-of-the-art method OW-DETR, our SS-OWFormer with merely $10\%$  achieves an absolute gain of up to $5\%$ in unknown recall for task-2 and task-3 owing to the object query-guided pseudo-labeling. 
Furthermore, our SS-OWFormer with just $10\%$ labeled data outperforms state-of-the-art OW-DETR trained with $50\%$ labeled data in terms of mAP for all the tasks, while SS-OWFormer with $50\%$ labeled data stands comparable to fully supervised state-of-the-art OW-DETR. 
This poses SS-OWFormer as closer to a realistic solution by overcoming the requirement of fully supervised incremental learning for the OWOD problem. 

\subsection{Ablation Studies}
Tab. \ref{tab:ablation} shows improvements made in performance with different components. Just using encoder features for pseudo-labeling instead of backbone features gives a $2\%$ improvement in Unknown Recall over the baseline.
Our proposed object query-guided pseudo-labeling helps to gain another $4.2\%$ in Unknown Recall and reach $12.26$ by utilizing decoder queries finally providing a relative gain of nearly doubling over the baseline in terms of Unknown Recall. 
Other design choice trials for the pseudo-labeling scheme include taking the mean of and maximum among the objectness scores. Mean is currently used in the proposed SS-OWFormer framework, while taking maximum causes a slight drop to unknown recall as it falls to $11.25$. 

\begin{table}[t]
\centering
\begin{tabular}{|c|c|c|c|c|c|c|}
\hline
\textbf{CJ}           & \textbf{GB}           & \textbf{GR}           & \textbf{PS}           & \textbf{SL}           & \multicolumn{1}{c|}{\textbf{U-Recall}} & \textbf{mAP} \\ \hline \hline
\xmark & \xmark & \xmark & \cmark & \cmark & 9.57                                   & 37.68        \\
\cmark & \cmark & \cmark & \cmark & \cmark & 10.03                                  & 38.13        \\
\cmark & \xmark & \xmark & \xmark & \xmark & 10.06                                  & 38.71        \\
\xmark & \cmark & \cmark & \xmark & \xmark & 10.11                                  & 38.89        \\
\cmark & \cmark & \cmark & \xmark & \xmark & 10.56                                  & 39.20        \\ \hline
\end{tabular}
\caption{Performance comparison when using different combinations of augmentation techniques applied together. The augmentations are abbreviated as CJ - Color Jitter, GB - Gaussian Blur, GR - Greyscale, PS - Posterize, and SL - Solarize. All the experiments are run using a fixed seed of 42 for Task 2 50\% labeled  data.}
\label{tab: aug_ablation}
\end{table}

\subsection{Experiments on Satellite Images} \label{subsec: exp_sat}
Tab. \ref{tab:sat1} reports unknown recall (U-recall) for task 1 supervised training which assesses the model's capability to capture unknown objects for the OWOD-S split.
The qualitative results for satellite images can be seen in Fig.\ref{fig:sat_viz}.
Our baseline achieves an unknown recall of $2.5$ on Task-1, with an mAP of $64.9$ on our OWOD-S task-1 benchmark. On the same task, SS-OWFormer achieves an unknown recall of $7.6$ and $66.7$ mAP and an mAP of $70.9$ for task-2. 
The object query-guided pseudo-labeling scheme feeds into the novelty classification and objectness branches which helps build a better separation of unknown objects from knowns and background in the satellite images.
Tab.\ref{tab:sat2} shows a comprehensive comparison of  our semi-supervised incremental learning strategy at different proportions of labeled and unlabeled data with a steady improvement over baseline under similar settings. This consistent improvement shown under the limited annotation availability setting emphasizes the importance of proposed contributions in a close to realistic satellite OWOD scenario without drastic forgetting of previously known classes. 
\begin{table*}[!t]
\centering
\scalebox{1}{
\begin{tabular}{|l|cc|cc|cc|cc|c|}
\hline
\multirow{2}{*}{\textbf{Method}} & \multicolumn{2}{c|}{\textbf{Task 1}} & \multicolumn{2}{c|}{\textbf{Task 2}} & \multicolumn{2}{c|}{\textbf{Task 3}} & \multicolumn{2}{c|}{\textbf{Task 4}} & \textbf{Task 5} \\ \cline{2-10} 
 & \multicolumn{1}{c|}{\textbf{UR}} & \textbf{mAP} & \multicolumn{1}{c|}{\textbf{UR}} & \textbf{mAP} & \multicolumn{1}{c|}{\textbf{UR}} & \textbf{mAP} & \multicolumn{1}{c|}{\textbf{UR}} & \textbf{mAP} & \textbf{mAP} \\ \hline
\textbf{OW-DETR} & \multicolumn{1}{c|}{13.6} & 21.2 & \multicolumn{1}{c|}{17.8} & 18.6 & \multicolumn{1}{c|}{12.7} & 16.7 & \multicolumn{1}{c|}{17.8} & 16.3 & 15.8 \\ \hline
\textbf{Ours (50\%)} & \multicolumn{1}{c|}{16.7} & 23.3 & \multicolumn{1}{c|}{21.6} & 18.1 & \multicolumn{1}{c|}{14.9} & 15.9 & \multicolumn{1}{c|}{18.7} & 15.4 & 15.1 \\ \hline
\end{tabular}%
}
\caption{Comparison over the proposed OWOD splits on Objects365 dataset. Our splits derived from a subset of Objects365, comprise 100k images and five different tasks. Task-1 has 85 categories, while Tasks-2:5 have 70 categories each.  To our knowledge, we are the first to report OWOD results on  Objects365, and our method performs favorably compared to OW-DETR \cite{gupta2021ow}}
\label{tab: new_Object365}
\end{table*}

\begin{table}[!ht]
\centering
\scalebox{1}{
\begin{tabular}{|l|c|c|c|}
\hline
\textbf{Method} & \textbf{\begin{tabular}[c]{@{}c@{}}Avg of 10 \\ Base classes\end{tabular}} & \textbf{\begin{tabular}[c]{@{}c@{}}Avg of 10 \\ Novel classes\end{tabular}} & \textbf{mAP} \\ \hline
\textbf{ORE} & 60.37 & 68.79 & 64.5 \\ \hline
\textbf{OW-DETR} & 63.48 & 67.88 & 65.7 \\ \hline
\textbf{Ours (50\%)} & 63.85 & 66.53 & 65.2 \\ \hline
\end{tabular}
}
\caption{Incremental detection results on PASCAL VOC (10+10 setting) as in \cite{joseph2021open} averaged over  base, novel classes \& overall mAP.}
\label{tab: incremental}
\end{table}

\subsection{Augmentation Techniques}
Tab.~\ref{tab: aug_ablation} shows performance comparison when using different combinations of augmentation techniques. We use color jitter, gaussian blur, random greyscale, posterizing, and solarizing as augmentations for the semi-supervised open-world learning pipeline.
From our experiments, we observe that posterizing and solarizing degrade the overall performance of the model as they are not well suited for \mbox{our problem setting.}

\subsection{Incremental Object Detection}
As shown in Tab.~\ref{tab: incremental} our SS-OWFormer performs favourably compared to previous works on incremental object detection (iOD) task. iOD experiments are performed on Pascal VOC \cite{everingham2010pascal} benchmark on the 10 + 10 class setting as proposed in \cite{joseph2021open}. Our SS-OWFormer achieves $65.2$ mAP while using only 50\% labeled data in the incremental learning setting compared with ORE and OW-DETR.

\subsection{Objects365 Benchmark}
Tab.~\ref{tab: new_Object365} compares our SS-OWFormer with OW-DETR on the Objects365 \cite{shao2019objects365} benchmark. Experiments on Objects365 with 365 object categories learned incrementally in a semi-supervised manner shows the effectiveness of our approach in a close to realistic setting. Our method consistently improves over the baseline OW-DETR while using 50\% labeled data and to the best our knowledge we are the first to report results on Objects365 in the Open-world object detection paradigm. 

%% file: sec/5_relation.tex
\section{Relation to Prior Art}
Semi-supervised and incremental object detection have been active research areas in computer vision, and recent works have achieved promising results. For semi-supervised object detection, methods such as S$^4$L~\cite{Zhai_2019_ICCV} and FixMatch~\cite{Sohn_2020_FixMatch} have been proposed to leverage unlabeled data by exploring consistency regularization techniques. S$^4$L incorporates self-supervised learning with semi-supervised learning and achieved state-of-the-art performance on various datasets. FixMatch utilizes a mix of labeled and unlabeled data, achieving competitive results with fully supervised approaches. Other approaches like \cite{LI_Learning_2019, Ziyue_Dual} introduce meta-learning to further enhance performance. For incremental object detection, methods like COCO-FUNIT~\cite{Saito_coco_FUNIT}, iCaRL~\cite{Rebuffi_2017_CVPR}, and NCM~\cite{Ristin_2014_CVPR} have been proposed to incrementally update the object detector model with new classes. COCO-FUNIT utilizes domain adaptation techniques for incremental learning, while iCaRL and NCM utilize exemplar-based methods for incremental feature learning. Other approaches like \cite{Kirsch_BatchBALD, Sinha_2019_ICCV, Yoo_2019_CVPR} utilize active learning and discriminative features to \mbox{further enhance performance.}

Open-world object detection in natural images recently gained popularity due to its applicability in real-world scenarios. 
ORE~\cite{joseph2021open} introduces an open-world object detector based on the two-stage Faster R-CNN~\cite{ren2015faster}. Since unknown objects are not annotated for training in the open-world paradigm, ORE utilizes an auto-labeling step to obtain a set of pseudo-unknowns for training. 
The OW-DETR \cite{gupta2021ow} introduces an end-to-end transformer-based framework for open-world object detection with attention-driven pseudo-labeling, novelty classification, and an objectness branch to triumph over the OWOD challenges faced by ORE. Methods like \cite{Saito_coco_FUNIT,Rebuffi_2017_CVPR,Ristin_2014_CVPR,perez2020incremental} have been proposed to incrementally update the object detector model with new classes.
OW-DETR achieved state-of-the-art performance on open-world object detection on the MS COCO benchmark.  
Localizing objects in satellite imagery\cite{xia2018dota, waqas2019isaid, cheng2022anchor} is a challenging task\cite{aleissaee2022transformers, van2018you, gong2022swin}.
The state-of-the-art results on DOTA \cite{xia2018dota} dataset is achieved by \cite{wang2022advancing} by adapting the standard vision transformer to remote sensing domain using rotated window attention.
To the best of our knowledge, open-world object detection has been focused on natural images and we are the first to propose an open-world object detection \mbox{problem for satellite images.}

%% file: sec/6_conclusion.tex
\section{Conclusion}
We present SS-OWFormer, a framework aiming to reduce reliance on external oracles in the OWOD problem.
SS-OWFormer comprises object query-guided pseudo-labeling to overcome limitations faced by heuristic approaches followed in previous works. We further explore a semi-supervised open-world object detection framework and introduce an OWOD-S split on DOTA. Experiments reveal the benefits of our contributions, leading to improvements for both known and unknown classes. Lastly, we validate our contributions in natural and remote sensing domains, achieving state-of-the-art OWOD performance.

%% file: sec/7_ethics.tex
\section*{Ethics Statement}
In alignment with the AAAI Ethics Policy, we address the ethical dimensions of our work on Semi-Supervised Open-World Object Detection. We have conscientiously credited the data sources and other open source works on which SS-OWFormer is built upon.The open-world object detection problem is an intriguing real-world scenario that gradually learns additional objects. However, there may be circumstances in which a certain object or fine-grained category must not be identified because of privacy or legal issues, whether in satellite images or otherwise. 
Moreover, although  the proposed SS-OWFormer can incrementally learn new object categories, it does  not have an explicit mechanism to  forget some of the previously seen categories. Developing open-world object detectors with explicit forgetting mechanisms  will be an interesting future research direction. Our commitment to transparency is evident through the availability of open-source resources, and we value collaboration and accountability within the research community. In recognizing the broader societal impact of our research, we pledge to uphold ethical standards in the development and deployment of our model and its applications.

\section*{Acknowledgements}
The computational resources were provided by the National Academic Infrastructure for Supercomputing in Sweden (NAISS), partially funded by the Swedish Research Council through grant agreement no. 2022-06725, and by the Berzelius resource, provided by the Knut and Alice Wallenberg Foundation at the \mbox{National Supercomputer Centre.}

%% file: sec/X_suppl.tex
\clearpage

\setcounter{page}{1}
\maketitlesupplementary

\maketitle
\section{Proposed satellite OWOD Splits }

Open-world object detection is a challenging paradigm because of the open structure of the problem.  In this work, we bring open-world object detection for the first time to the field of satellite imagery.
For satellite OWOD, we adapt our baseline for natural images by modifying the regression branch to predict oriented boxes.
To facilitate research in this  challenging problem setting, we introduce our novel  satellite OWOD (OWOD-S) split. 
This data split proposed in  our work tries to avoid data leakage across tasks and ensures fair usage of the entire data at the same time. 
Similar object classes like small and large vehicles are put in the same task to  avoid data leakage.

Each task denoted by a percentage amount is prepared such that it contains the mentioned proportion of annotated instances from each of the currently known object classes.  Making splits based on object instances instead of the number of images produces a fairer distribution among all object categories because randomly dividing by images may create an unfair concentration of some classes that occur in high density. 
For example, storage tanks (up to 427 instances in an image) are seen to occur in much higher density as compared to roundabouts (up to 14 instances in an image).

Table \ref{tab:split_instances} shows the proposed Split-1 and Split-2 for open-world satellite object detection. For both splits,  we show the classes introduced in each task. \\
\noindent\textbf{Split-1:} The Split-1  comprises two tasks: \textit{task-1} with small- and large- vehicles, ships, planes, helicopters, harbors, swimming pools, ground track fields, tennis courts while \textit{task-2} consists of soccer ball fields, basketball courts, baseball diamonds, bridges, roundabouts, and storage tanks.  
\noindent\textbf{Split-2:} Split-2 divides the object classes into four tasks. This is done to imitate a realistic setting where a model is required to learn a few new categories when necessary or needed instead of a large pool of object classes to be learned incrementally. Here \textit{task-1} consists of small- and large- vehicles, ships, planes, helicopters, harbors, swimming pools, ground track fields, and tennis courts;  \textit{task-2} comprises basketball courts and baseball diamonds,  \textit{task-3} with bridges and roundabouts, and the final task, \textit{task-4}, has storage tanks and soccer ball fields.

\begin{table}[htb!]
\centering
\small
\scalebox{0.96}{
\begin{tabular}{|c|c|c|c|c|}
\hline
\textbf{Splits}            & \textbf{Task} & \textbf{Classes}                                                                       & \textbf{Images} & \textbf{Instances} \\ \hline \hline
\multirow{6}{*}{Split1} & Task 1        & \begin{tabular}[c]{@{}l@{}}SV, LV, SH, \\ PL, HC, HA, \\ SP, GTF, TC\end{tabular}      & 32711           & 448443             \\ \cline{2-5} 
                           & Task 2 100\%  & \multirow{3}{*}{\begin{tabular}[c]{@{}l@{}}SBF, BC, BD,\\ BR, RA, ST\end{tabular}} & 19547           & 346482             \\
                           & Task 2 75\%   &                                                                                        & 14721           & 127192             \\
                           & Task 2 50\%   &                                                                                        & 8626            & 75771              \\
                           \hline
\multirow{7}{*}{Split2} & Task 1        & \begin{tabular}[c]{@{}l@{}}SV, LV, SH, \\ PL, HC, HA, \\ SP, GTF, TC\end{tabular}      & 32711           & 448443             \\ \cline{2-5} 
                           & Task 2 100\%  & \multirow{2}{*}{BC, BD}                                                                & 2363            & 10550              \\
                           & Task 2 50\%   &                                                                                        & 1226            & 5403               \\ \cline{2-5} 
                           & Task 3 100\%  & \multirow{2}{*}{BR, RA}                                                                & 10550           & 42789              \\
                           & Task 3 50\%   &                                                                                        & 5238            & 22587              \\ \cline{2-5} 
                           & Task 4 100\%  & \multirow{2}{*}{ST, SBF}                                                               & 6634            & 110855             \\
                           & Task 4 50\%   &                                                                                        & 3283            & 56667              \\ \hline
                           & Test set      &                                                                                        & 5634            & 79987              \\ \hline
\end{tabular}
}
\caption{\textbf{Task composition in the proposed satellite OWOD (OWOD-S) splits for open-world evaluation in satellite images.} In this work, we introduce two splits named Split-1 and Split-2 dividing the object classes into two and four incremental learning tasks, respectively.  
In  both splits, the object classes for each task are shown along with the total number of images and object instances.  
Each partially labeled task denoted by a percentage amount is prepared such that it contains  the mentioned proportion of annotated instances from each of the currently known object classes. 
Test sets for each task are prepared such that all classes introduced till that task have class labels while others remain annotated as \textit{unknown}. The object classes are abbreviated as \textit{SV - small-vehicle, LV - large vehicle, SH - ship, PL - plane, HC - helicopter, HA - harbor, SP - swimming pool, GTF - ground track field, TC - tennis court, SBF - soccer ball field, BC - basketball court, BD - baseball diamond, BR - bridge, RA - roundabout, ST - storage tank.}}
\label{tab:split_instances}
\end{table}

\begin{figure*}[ht]
    \centering
    \includegraphics[scale=0.6]{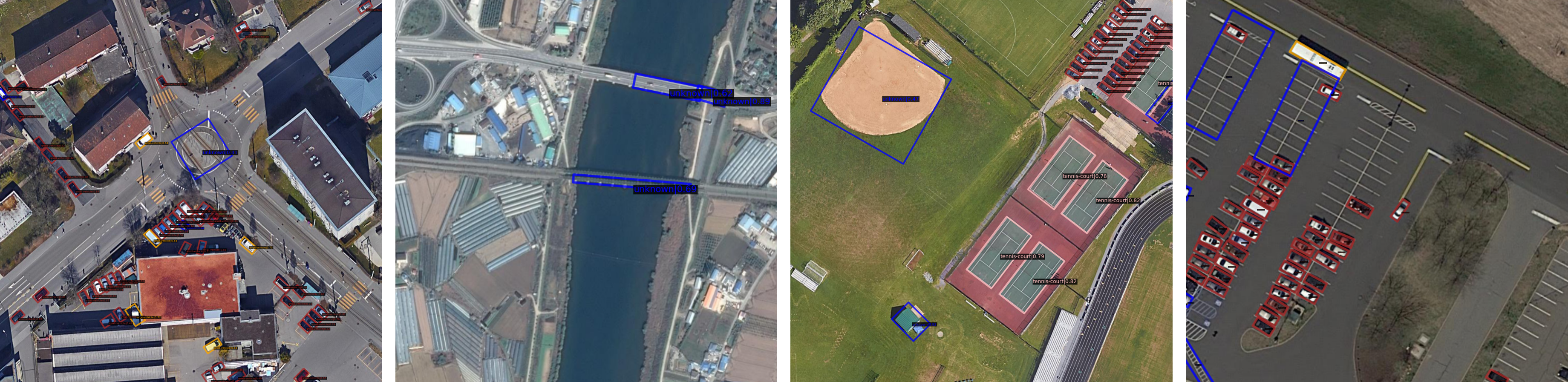}
    \setlength{\belowcaptionskip}{5pt}
    \caption{\textbf{Qualitative results showing the detections in satellite images.} Bounding boxes are illustrating known and unknown (\textcolor{blue}{blue}) objects predicted by our SS-OWFormer.
    The first example shows the known category \textit{small-vehicle} being detected with unknown category \textit{roundabout} and the next examples respectively show that the unknown categories \textit{bridges} and \textit{baseball diamond} are being detected.
    In the last example, although an empty parking lot is not a valid object category of the dataset, the model `wrongly' predicts it  as an unknown object. We conjuncture that this is likely due to the visual similarity of an empty parking lot  with the \textit{harbor} category. 
}
    \label{fig:base_vs_rappl}
\end{figure*}

\begin{figure*}[!ht]
    \centering
    \includegraphics[scale = 0.87]{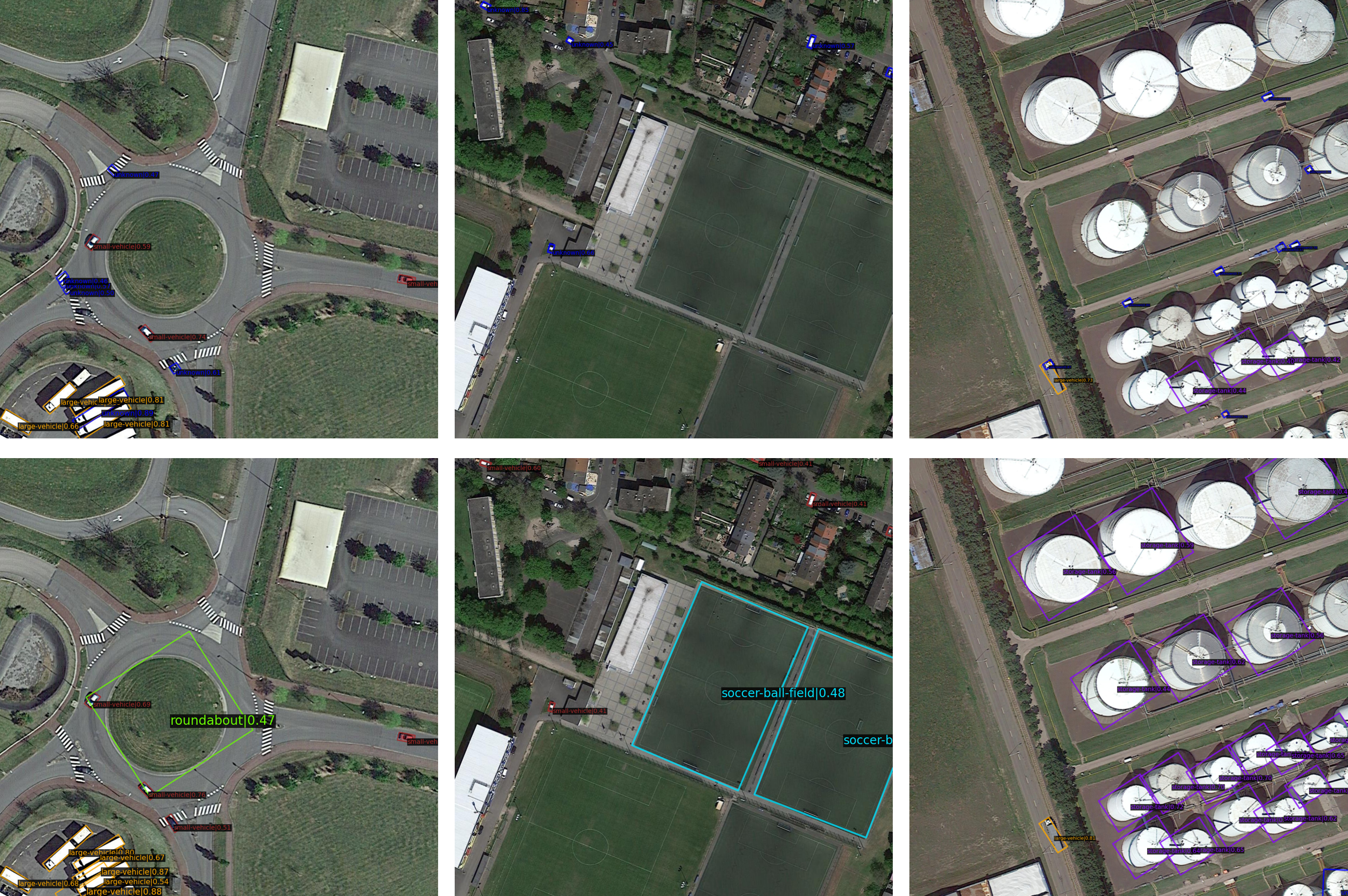}
    \caption{\textbf{Qualitative results  without (top row) and  with (bottom row) our proposed semi-supervised learning strategy for 50\% labeled satellite OWOD Split-1 Task-2 data.} Bounding boxes are illustrating known and unknown (\textcolor{blue}{blue}) objects.
   Examples 1 and 2 show that our method is  able to accurately detect (localize and recognize)  newly introduced categories of \textit{roundabout} and \textit{soccer ball field},  without forgetting the previously introduced classes such as \textit{small vehicle} and \textit{large vehicle}. 
   Moreover, our SS-OWFormer is able to detect additional instances of the newly introduced category of (\textit{storage tank})  in example 3. (best viewed in zoom) }
    \label{fig:base_vs_ssl25}
\end{figure*}

\begin{table*}[!ht]
\centering
\scriptsize
\begin{tabular}{|c|cc|cccc|cccc|ccc|}

\hline
\multicolumn{1}{|c|}{\textbf{Task IDs}} & \multicolumn{2}{c|}{\textbf{Task-1}}                                                                                               & \multicolumn{4}{c|}{\textbf{Task-2 }}                                                                                                                                                                                   & \multicolumn{4}{c|}{\textbf{Task-3}}                                                                                                                                                                                   & \multicolumn{3}{c|}{\textbf{Task-4}}                                                                                                                      \\ \hline
              \textbf{Method}                         &                                    & \textbf{mAP}                                 &                                     & \multicolumn{3}{c|}{\textbf{mAP}}                                                                                                &                                     & \multicolumn{3}{c|}{\textbf{mAP}}                                                                                                & \multicolumn{3}{c|}{\textbf{mAP}}                                                                                                 \\
\multirow{-2}{*}{\textbf{}}            & \multirow{-2}{*}{\textbf{U-Recall}} & \textbf{\begin{tabular}[c]{@{}c@{}}Prev\end{tabular}} & \multirow{-2}{*}{ }\textbf{U-Recall} & \textbf{\begin{tabular}[c]{@{}c@{}}Prev\end{tabular}} & \textbf{\begin{tabular}[c]{@{}c@{}}Cur\end{tabular}} & \textbf{Both} & \multirow{-2}{*}{ }\textbf{U-Recall} & \textbf{\begin{tabular}[c]{@{}c@{}}Prev\end{tabular}} & \textbf{\begin{tabular}[c]{@{}c@{}}Cur\end{tabular}} & \textbf{Both} & \textbf{\begin{tabular}[c]{@{}c@{}}Prev\end{tabular}} & \textbf{\begin{tabular}[c]{@{}c@{}}Cur\end{tabular}} & \textbf{Both} \\ \hline \hline
Baseline (100\%)                              &  2.5                                 & 64.9                                                                 &  2.8                                 & 73.6                                                                 & 44.9                                                              & 63.4          &  2.9                                 & 64.9                                                                 & 29.7                                                              & 56            & 62.4                                                                 & 52.2                                                              & 61.9          \\
\textbf{SS-OWFormer}  (50\%)                    &  \textbf{7.6}                                   & 66.7                                                                    &  \textbf{6.7}                                 & 73.5                                                                 & 41.2                                                              & 59.8          &  \textbf{8.3}                                 & 64.1                                                                 & 42.2                                                              & 56.4          & 60.1                                                                 & 52.4                                                              & 59.1          \\ \hline
\end{tabular}
\caption{\textbf{Comparison between baseline and SS-OWFormer on proposed satellite OWOD Split-2.} The comparison is shown in terms of mAP for known classes and unknown class recall (U-Recall). The unknown recall metric assesses the model's ability to capture unknown objects. Our SS-OWFormer using only 50\% annotated  data during tasks 2-4 is achieving   a performance (both U-Recall and mAP)  comparable with the baseline requiring 100\% annotated data for all four tasks. 
}
\label{tab: split2results}
\end{table*}

\begin{table*}[ht]
\centering
\scriptsize
\begin{tabular}{|l|cc|cccc|cccc|ccc|}
\hline
\multicolumn{1}{|l|}{\multirow{3}{*}{\textbf{Method}}} & \multicolumn{2}{c|}{\textbf{Task1}}                                                     & \multicolumn{4}{c|}{\textbf{Task2 (50\%)}}                                                                                                              & \multicolumn{4}{c|}{\textbf{Task3 (50\%)}}                                                                                                              & \multicolumn{3}{c|}{\textbf{Task4 (50\%)}}                                                    \\ \cline{2-14} 
\multicolumn{1}{|c|}{}                                 & \multicolumn{1}{c|}{\multirow{2}{*}{\textbf{U-Recall}}} & \multirow{2}{*}{\textbf{mAP}} & \multicolumn{1}{c|}{\multirow{2}{*}{\textbf{U-Recall}}} & \multicolumn{3}{c|}{\textbf{mAP}}                                                      & \multicolumn{1}{c|}{\multirow{2}{*}{\textbf{U-Recall}}} & \multicolumn{3}{c|}{\textbf{mAP}}                                                      & \multicolumn{3}{c|}{\textbf{mAP}}                                                      \\
\multicolumn{1}{|c|}{}                                 & \multicolumn{1}{c|}{}                                   &                               & \multicolumn{1}{c|}{}                                   & \multicolumn{1}{c|}{\textbf{Prev}} & \multicolumn{1}{c|}{\textbf{Cur}} & \textbf{Both} & \multicolumn{1}{c|}{}                                   & \multicolumn{1}{c|}{\textbf{Prev}} & \multicolumn{1}{c|}{\textbf{Cur}} & \textbf{Both} & \multicolumn{1}{c|}{\textbf{Prev}} & \multicolumn{1}{c|}{\textbf{Cur}} & \textbf{Both} \\ \hline \hline
OW-DETR (T1 50\%)                                & \multicolumn{1}{c|}{6.09}                               & 52.78                         & \multicolumn{1}{c|}{6.82}                               & \multicolumn{1}{c|}{47.31}         & \multicolumn{1}{c|}{17.16}        & 32.23         & \multicolumn{1}{c|}{7.13}                               & \multicolumn{1}{c|}{30.2}          & \multicolumn{1}{c|}{8.13}         & 22.84         & \multicolumn{1}{c|}{22.12}         & \multicolumn{1}{c|}{5.63}         & 17.99         \\
\textbf{SS-OWFormer (T1 50\%)}                                   & \multicolumn{1}{c|}{11.13}                              & 55.51                         & \multicolumn{1}{c|}{10.36}                              & \multicolumn{1}{c|}{48.32}         & \multicolumn{1}{c|}{24.79}        & 36.55         & \multicolumn{1}{c|}{12.37}                              & \multicolumn{1}{c|}{35.98}         & \multicolumn{1}{c|}{14.8}         & 28.92         & \multicolumn{1}{c|}{27.89}         & \multicolumn{1}{c|}{10.76}        & 23.6          \\ \hline
OW-DETR (T1 100\%)                           & \multicolumn{1}{c|}{6.17}                               & 58.53                         & \multicolumn{1}{c|}{6.94}                               & \multicolumn{1}{c|}{50.53}         & \multicolumn{1}{c|}{19.28}        & 34.91         & \multicolumn{1}{c|}{7.64}                               & \multicolumn{1}{c|}{32.7}          & \multicolumn{1}{c|}{9.13}         & 24.85         & \multicolumn{1}{c|}{24.08}         & \multicolumn{1}{c|}{5.74}         & 19.49         \\
\textbf{SS-OWFormer (T1 100\%)}                               & \multicolumn{1}{c|}{12.26}                              & 59.85                         & \multicolumn{1}{c|}{10.56}                              & \multicolumn{1}{c|}{52.04}         & \multicolumn{1}{c|}{26.35}        & 39.2          & \multicolumn{1}{c|}{13.16}                              & \multicolumn{1}{c|}{39.46}         & \multicolumn{1}{c|}{13.63}        & 30.85         & \multicolumn{1}{c|}{29.97}         & \multicolumn{1}{c|}{11.48}        & 25.35         \\ \hline
\end{tabular}
\caption{\textbf{Effects of using the semi-supervised setting for task-1 on performance over the learning cycle.}The comparison is presented in terms of unknown recall (U-Recall) and the previously known (Prev), current known (Cur), and Overall (both) AP for all tasks. U-Recall is not reported for task-4 since all classes are known. 
The first two rows show task-1 utilizing semi-supervised learning with 50\% labeled data, while the next two rows have a fully supervised task-1. 50\% labeled and 50\% unlabeled data is used for tasks-2,3 and 4 in all cases.
} 
\label{tab: semi_sup_task1}
\end{table*}

\section{Results on satellite OWOD Splits}
We evaluate the performance of our SS-OWFormer on both Split-1 and Split-2 proposed for satellite images. Here, the test set for both splits is prepared from a common pool of images shared across all tasks. The test image annotations for each task are prepared such that all known (previously and currently introduced) classes have their corresponding class labels while others remain annotated as \textit{unknown}. 
The object classes to be introduced in future tasks are used to evaluate unknown detections based on Unknown Recall (U-Recall) metric and known classes are  evaluated using mAP.

Tab.~3 in the main paper and Tab.~\ref{tab: split2results} here show that our SS-OWFormer maintains similar performance on split-1 and split-2 irrespective of the number of incremental learning tasks. 
This also shows that SS-OWFormer is robust to drastic forgetting of previously known categories under both fully-supervised and semi-supervised open-world detection settings, even when there are more subsequent tasks.

\noindent\textbf{Qualitative results:} 
Fig.~\ref{fig:base_vs_ssl25} shows a comparison between the baseline model and SS-OWFormer on satellite images  when both are trained with 50\% labeled data where SS-OWFormer is trained with the proposed semi-supervised learning setup. 
For each example, baseline results are shown at the top, while predictions of SS-OWFormer are shown at the bottom.
The results show that the  SS-OWFormer with the proposed object query-guided pseudo-labeling can better detect  unknowns in comparison to  the baseline. 
In the first two examples, SS-OWFormer detects fewer false unknowns and correctly detects known classes of \textit{roundabout} and \textit{soccer ball fields} in the subsequent task.
The third example shows SS-OWFormer being able to detect more number of \textit{storage tanks} compared to the baseline.

\begin{table}[t]
\centering
\begin{tabular}{|l|cc|}
\hline
\textbf{Method}         & \textbf{U-Recall} & \textbf{mAP} \\ \hline \hline
ORE-EBUI       & 4.9               & 56           \\
OW-DETR        & 7.5               & 59.2         \\
\textbf{SS-OWFormer}           & 12.26             & 59.85        \\ \hline
OW-DETR (50\%) & 7.33              & 52.78         \\
\textbf{SS-OWFormer (50\%)}    & 11.13             & 55.51        \\ \hline
OW-DETR (25\%) & 7.99              & 42.86        \\
\textbf{SS-OWFormer (25\%)}    & 11.98             & 46.87        \\ \hline
OW-DETR (10\%) & 8.1               & 29.75        \\
\textbf{SS-OWFormer  (10\%)}   & 11.92             & 32.81        \\ \hline
\end{tabular}
\caption{\textbf{Comparison of performance of our SS-OWFormer with previous works on Task-1 OWOD~\cite{joseph2021open} splits .} The comparison is made for 4 proportions of labeled and unlabeled data, with the percentage of labeled data indicated in parentheses.}
\label{tab: task1_comparison}
\end{table}

\begin{figure}[!ht]
    \centering
    \includegraphics[width=0.47\textwidth]{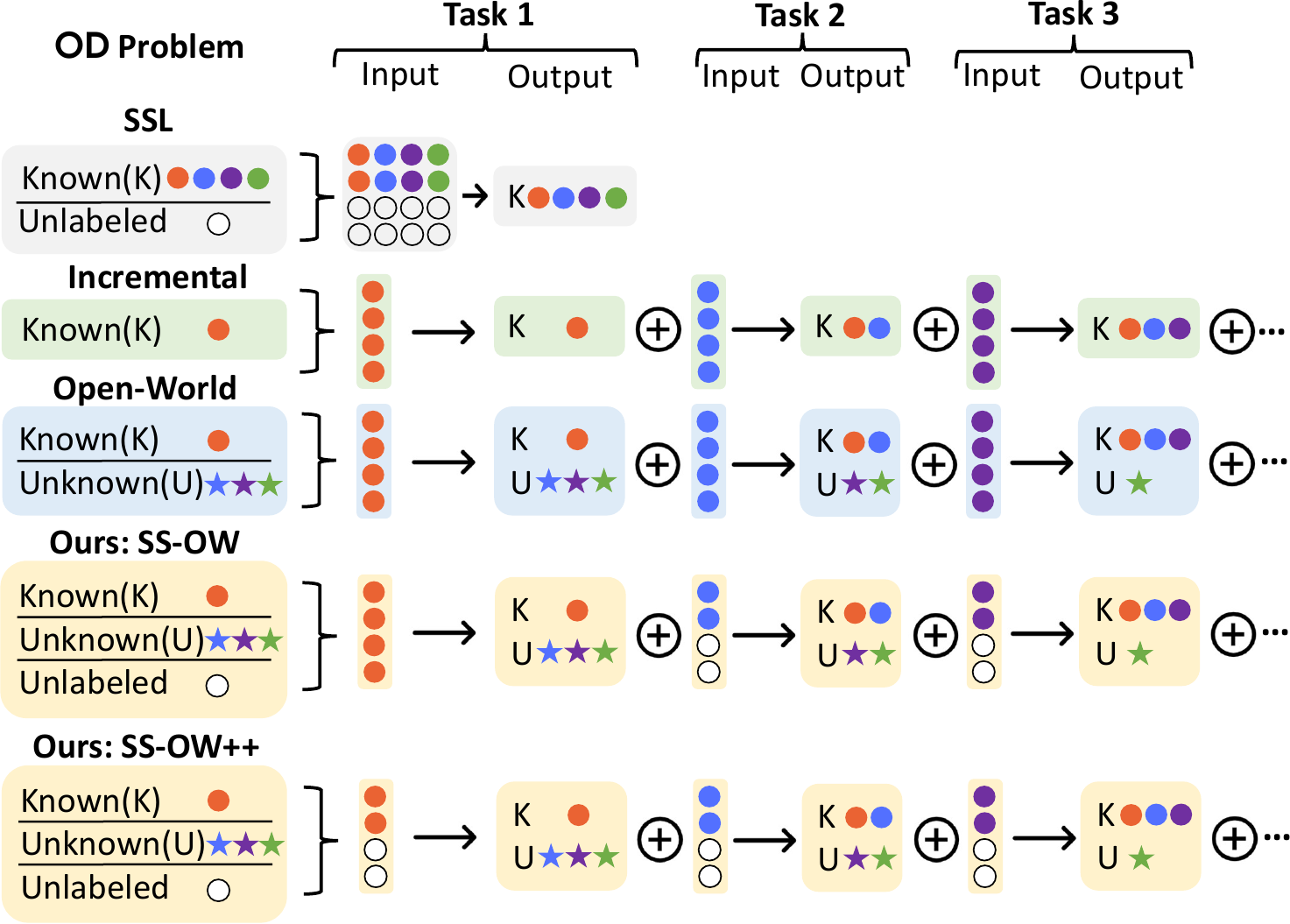}
    \caption{\textbf{Comparison of our SS-OW and special case SS-OW++ with other closely related object detection problem settings.} The special case SS-OW++ depicted in the last row shows how the first task or the model initialization can be done with a mix of labeled and unlabeled data just like the subsequent tasks. The effects of this setup on performance can be seen in Tab.~\ref{tab: semi_sup_task1}.}
    \label{fig: problem_figure_special}
\end{figure}

\section{Additional Experiments}
\subsection{SS-OWOD++: SS for \textit{all}  Tasks}
The proposed SS-OWOD++ is a special case of SS-OWOD (see Fig.~\ref{fig: problem_figure_special}) 
when the first task (Task-1) in the SS-OWOD is also  redesigned to use both labeled and unlabeled data instead of being fully supervised (i.e. all four tasks including Task-1 are semi-supervised). It is observed that the effects of doing so do not result in drastic drops in performance \mbox{as seen in Tab.~\ref{tab: semi_sup_task1}.} 

Tab.~\ref{tab: method_compare} compares our work with other prominent semi-supervised (SS) and weakly-supervised (WSOD) methods. Previous works like \cite{yang2019detecting,redmon2017yolo9000,gao2019note} can detect objects in an SS or WSOD manner. However these methods cannot explicitly detect unknown objects or incrementally learn new classes when data becomes available. Previous OWOD works \cite{joseph2021open, gupta2021ow} also employ  pseudo-labeling  based on naive heuristics such as such simple averaging or clustering. In contrast, our SS-OWFormer tries to learn objectness  using properties of transformer encoder \textit{and} decoder  to pseudo-label \textit{unknown} objects \textit{never seen} by the model, while also operating in a semi-supervised OWOD setting. 

\begin{table}[t]
\small
\centering
\scalebox{0.78}{
\begin{tabular}{|l|
>{\columncolor[HTML]{FFECEB}}c |
>{\columncolor[HTML]{FFECEB}}c |
>{\columncolor[HTML]{FFECEB}}c |
>{\columncolor[HTML]{FFFFC7}}c |
>{\columncolor[HTML]{FFFFC7}}c |
>{\columncolor[HTML]{DFFFDE}}c |}
\hline
Methods             & \begin{tabular}[c]{@{}c@{}}Detecting\\ 11k classes\end{tabular} & \begin{tabular}[c]{@{}l@{}}Yolo\\ 9000\end{tabular} & \begin{tabular}[c]{@{}c@{}}Note\\ RCNN\end{tabular} & ORE   & \begin{tabular}[c]{@{}l@{}}OW-\\ DETR\end{tabular} & Ours  \\
\hline
Detect unknowns     & \xmark                                                           & \xmark    & \xmark     & \cmark & \cmark   & \cmark      \\ \hline
Non-stationary data & \xmark                                                           & \xmark    & \xmark     & \cmark & \cmark   & \cmark      \\ \hline
Semi-supervised     & \cmark                                                           & \cmark    & \cmark     & \xmark & \xmark   & \cmark     \\ \hline
\end{tabular}}
\caption{\textbf{Comparison of different detection methods} (SS: col 1-3; OWOD: col 4:5, and Ours) based on their ability to detect unknown objects, handle non-stationary data (class incremental) samples, and semi-supervised (SS) learning. Our SS-OWFormer is designed to detect \textit{unknown
objects} and then \textit{incrementally} learning these ‘unknown’ objects when introduced with only \textit{few } labeled samples in the subsequent tasks. }
\label{tab: method_compare}
\end{table}


\begin{figure*}[t]
    \centering
    \includegraphics[scale = 0.8]{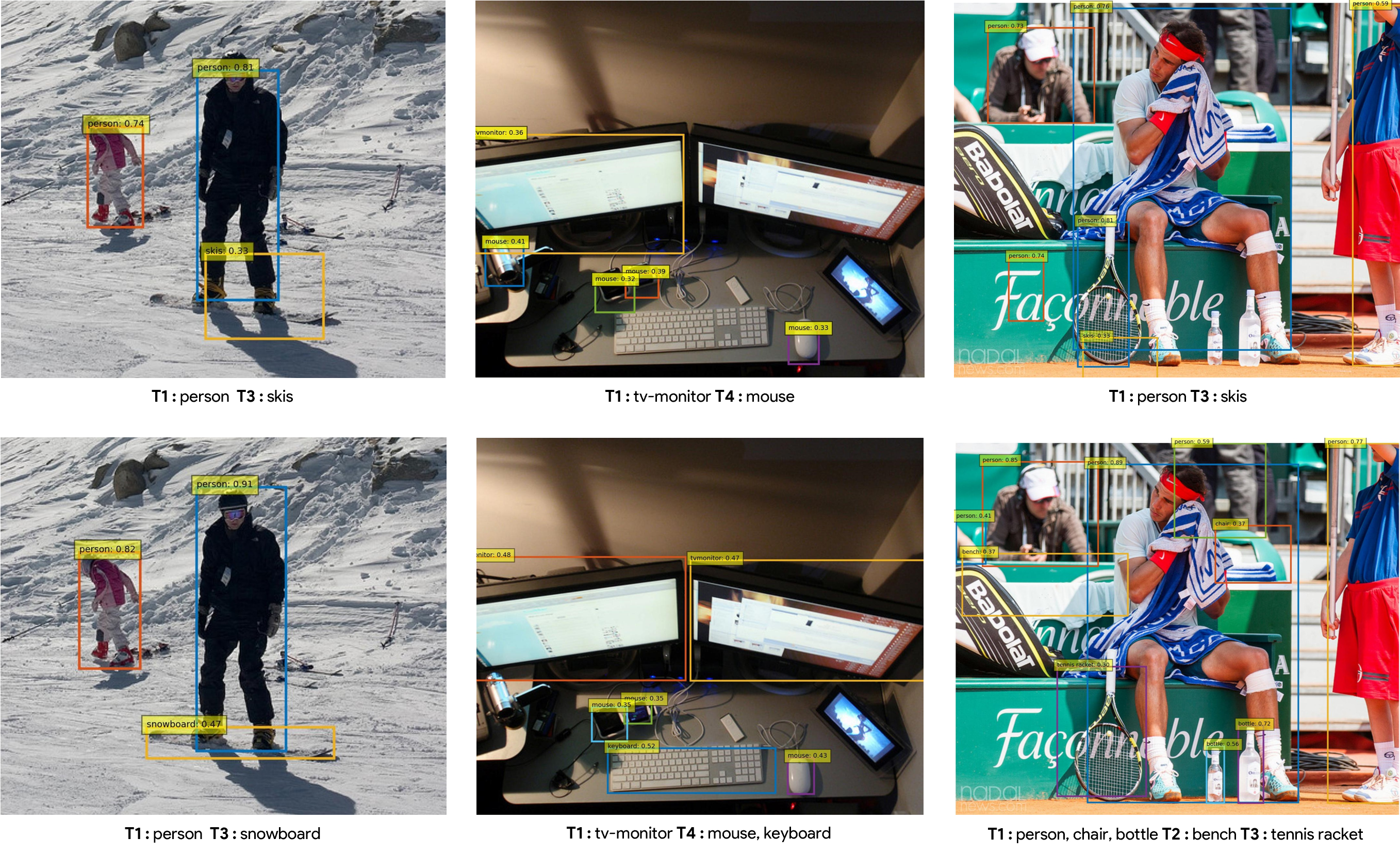}
    \caption{\textbf{Qualitative comparison between OW-DETR \cite{gupta2021ow} (top row) and our SS-OWFormer (bottom row) trained with 50\% of annotated data on MS COCO split.} In the first example, OW-DETR wrongly detects \textit{snowboard} as \textit{skis} while also being unable to localize properly whereas our SS-OWFormer  trained with 50\% of annotated data accurately detects \textit{snowboard} introduced in the previous tasks (task-3) with a tight bounding box. In the second example, OW-DETR misses detecting \textit{keyboard} and one instance of \textit{tv-monitor} while SS-OWFormer detects all instances of these objects. In the last example, OW-DETR obtains false detections of \textit{person} (task 1 category) on the scene text and fails to detect many object classes. Our SS-OWFormer accurately predicts the \textit{tennis racket} (task-3 category) without compromising on the detection performance of previously known categories such as  \textit{person, chair, bottle} (introduced in task-1) and \textit{bench} (introduced in task-2).}
    \label{fig:mscoco_qual}
\end{figure*}

\section{Societal Impacts and Limitations}
The open-world object detection problem is an intriguing real-world scenario that gradually learns additional objects. However, there may be circumstances in which a certain object or fine-grained category must not be identified because of privacy or legal issues, whether in satellite images or otherwise. 
Moreover, although  the proposed SS-OWFormer can incrementally learn new object categories, it does  not have an explicit mechanism to  forget some of the previously seen categories. Developing open-world object detectors with explicit forgetting mechanisms  will be an interesting \mbox{future research direction.}

In satellite imagery certain cases can occur like vehicles in parking lots or on bridges, where the foreground and background can both be objects of interest, leading to the increased complexity of  open-world object detection.   
The performances are nevertheless on the lower side as a result of the complexities of open-world object detection, aerial imagery along with partially annotated data.
We believe that this work will motivate future efforts in this difficult \mbox{yet realistic setting.}